\documentclass[10pt,twocolumn,letterpaper]{article}

\usepackage[pagenumbers]{cvpr} 

\usepackage{graphicx}
\usepackage{amsmath}
\usepackage{amssymb}
\usepackage{booktabs}
\usepackage{tabularx}
\usepackage{adjustbox}
\usepackage{array}
\usepackage{colortbl}
\usepackage{makecell}
\usepackage{float,wrapfig}
\usepackage{hhline}
\usepackage{multirow}
\usepackage{color} 
\usepackage{bbding}

\usepackage{amsthm}
\usepackage{mathrsfs}

\usepackage{diagbox}
\usepackage{subeqnarray}
\usepackage{cases}
\usepackage{setspace}

\usepackage{dcolumn}
\usepackage{float}
\usepackage{epsfig}

\usepackage{epstopdf}

\usepackage{xcolor}

\usepackage{bm}

\usepackage{overpic}
\usepackage{rotating}

\usepackage{enumitem}

\usepackage{pifont}
\newcommand{\cmark}{\ding{51}}%
\newcommand{\xmark}{\ding{55}}%
\usepackage[accsupp]{axessibility}  

\usepackage[pagebackref,breaklinks,colorlinks]{hyperref}

\def\cvprPaperID{6124} 
\def\confName{CVPR}
\def\confYear{2022}

\begin{document}

\title{
Retrieval-based Spatially Adaptive Normalization for Semantic Image Synthesis
\vspace{-4pt}}

\author{\small{Yupeng Shi $^{1}$, \ \ Xiao Liu $^{1}$, Yuxiang Wei $^{2}$, Zhongqin Wu $^{1}$, Wangmeng Zuo$^{2, 3}$ $^{(}$\Envelope$^)$ } \\ \vspace{-0.2em}
		$^1$\small{Tomorrow Advancing Life}, \ \ $^2$\small{Harbin Institute of Technology}, \ \ $^3$\small{Peng Cheng Laboratory}\\ \vspace{-0.2em}
		{\tt\small{ \{csypshi, ender.liux, yuxiang.wei.cs\}@gmail.com} } {\tt {\small{wuzhongqin@tal.com}} }  {\tt{\small{wmzuo@hit.edu.cn}}} \\
	}

\twocolumn[{%
\renewcommand\twocolumn[1][]{#1}%
\maketitle
\begin{center}
    \vspace{-10mm}
    \centering
    \captionsetup{type=figure}
    \includegraphics[width=0.99 \linewidth]{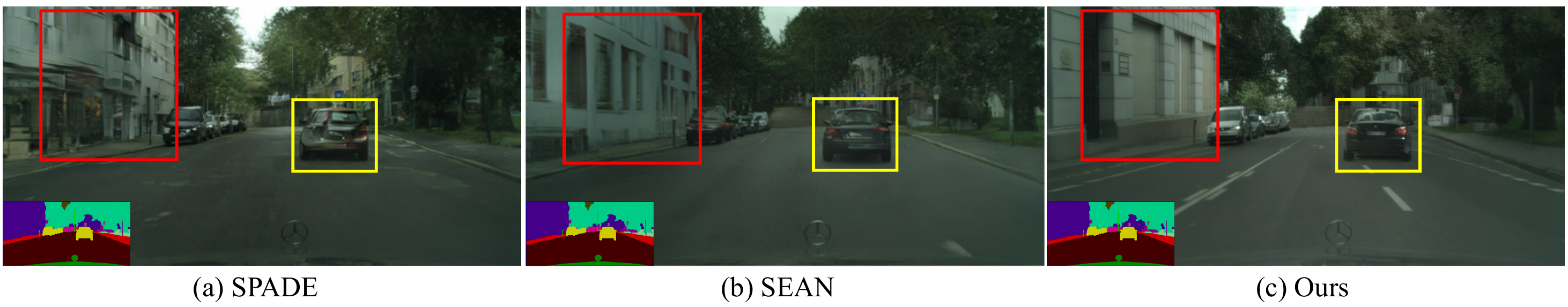}
    \vspace{-1em}
    \captionof{figure}{Synthesized results of SPADE~\cite{park2019semantic}, SEAN~\cite{zhu2020sean} and our method. (a) With the class-level guidance, SPADE produces blurry synthesis results. (b) With the region-level style vector, SEAN generates better details, but still prefers spatially uniform synthesis result. (c) Benefited from pixel level fine-grained guidance, our RESAIL is effective in generating visually plausible image with clear details. }
    \label{fig:cmp_spade_sean}
\end{center}%
\vspace{-0.77em}
}]

\begin{abstract}
\vspace{-1.0em}

Semantic image synthesis is a challenging task with many practical applications. Albeit remarkable progress has been made in semantic image synthesis with spatially-adaptive normalization, existing methods usually normalize the feature activations under the coarse-level guidance (e.g., semantic class). However, different parts of a semantic object (e.g., wheel and window of car) are quite different in structures and textures, making blurry synthesis results usually inevitable due to the missing of fine-grained guidance. In this paper, we propose a novel normalization module, termed as REtrieval-based Spatially AdaptIve normaLization (RESAIL), for introducing pixel level fine-grained guidance to the normalization architecture. {Specifically, we first present a retrieval paradigm by finding a content patch of the same semantic class from training set with the most similar shape to each test semantic mask.}
Then, the retrieved patches are composited into retrieval-based guidance, which can be used by RESAIL for pixel level fine-grained modulation on feature activations, thereby greatly mitigating blurry synthesis results.
Moreover, distorted ground-truth images are also utilized as alternatives of retrieval-based guidance for feature normalization, further benefiting model training and improving visual quality of generated images. 
Experiments on several challenging datasets show that our RESAIL performs favorably against state-of-the-arts in terms of quantitative metrics, visual quality, and subjective evaluation.
The source code is available at \url{https://github.com/Shi-Yupeng/RESAIL-For-SIS}. 
\vspace{-1em}
%
\end{abstract}

\vspace{-1em}
\section{Introduction}
\vspace{-0.2em}


Semantic image synthesis aims to generate photo-realistic image from the given semantic map. 
It is an important problem in computer vision that can be adopted in a variety of downstream tasks such as virtual idol, special effect, robotics~\cite{janai2020computer} and image manipulation~\cite{hong2018learning}. 

Humans have a remarkable ability to produce new creation from past experiences as references.  
In their early ages, children can paint a picture including flowers, sky and buildings by referring to templates of representative objects and backgrounds.
%
Thus, producing something from references is a natural way for image generation because editing the references and stitching them is relatively easy than creating the entire image out of thin air.
Inspired by this spirit, early works have well studied reference-based image synthesis, where proper references are searched from external memories \cite{hays2007scene,lalonde2007photo,chen2009sketch2photo,johnson2010cg2real,isola2013scene}.
Nonetheless, the retrieval, editing and stitching are conducted in separated and handcrafted manners, which are optimized in a sub-optimal way.
SIMS~\cite{qi2018semi} leverages deep network for further improving the quality of reference-based synthesized results, but it simply takes the retrieved image as network input, which is limited in synthesizing complex real-world scenes.
%

With the recent advance of deep generative networks, some recent studies \cite{park2019semantic, zhu2020sean, tan2021efficient, ntavelis2020sesame, schonfeld2021you} tackle semantic image synthesis using a spatially-adaptive normalization architecture, achieving significant performance improvements. 
However, with the coarse-level guidance (\eg, semantic class), these methods modulate the activations inside each semantic object in spatially uniform manner, regardless of the huge internal variation of the objects.
This inevitably leads to blurry results, especially for large  semantic object with complex parts. 
We take two representative spatially-adaptive normalization architectures as examples in Fig.~\ref{fig:cmp_spade_sean}. 
SPADE~\cite{park2019semantic} leverages the semantic layout as input and learns the modulation parameters through several convolution layers, being limited in generating high-quality object parts and leading to blurry synthesis results (Fig.~\ref{fig:cmp_spade_sean}(a)).
%
%
%
SEAN~\cite{zhu2020sean} improves SPADE by extracting style codes from selected regions, leading to flexible style control. 
However, the style map is generated by broadcasting the style codes to the corresponding semantic regions, which also prefers spatially uniform synthesis result  (Fig.~\ref{fig:cmp_spade_sean}(b)).
Most recent methods, \eg, CLADE~\cite{tan2021efficient} and OASIS~\cite{schonfeld2021you}, intrinsically are also based on coarse-level guidance.

In this paper, we tackle the above issues by presenting a novel feature normalization method, termed as REtrieval-based Spatially AdaptIve normaLization  (RESAIL).
Our intuition is two-fold. 
On the one hand, the object segment mask of the input semantic map can not only provide the semantic class but also the object shape. %
On the other hand, the training dataset contains rich shape and texture information of objects which cannot be entirely captured by the learned deep generative networks. 
%
%
%
%
Taking these intuitions into account, given a object segment mask, we present a retrieval paradigm for retrieving a segment image with the most similar shape from the training dataset. 
The retrieved segment images are then composited into a retrieval-based guidance, which naturally is spatially variant in pixel level.
%
%
%
We further propose a retrieval-based spatially adaptive normalization, where retrieval-based guidance and semantic map collaborate to provide pixel level fine-grained modulation on feature  activations. 
%
%
As shown in Fig.~\ref{fig:cmp_spade_sean}(c), benefited from pixel level fine-grained guidance, our RESAIL is effective in generating visually plausible image with clear details.
In contrast to SIMS~\cite{qi2018semi}, our method leverages retrieval-based guidance for spatially adaptive normalization, which is more effective in synthesizing photo-realistic images.
In comparison to SPADE~\cite{park2019semantic} and SEAN~\cite{park2019semantic}, our RESAIL can effectively leverage pixel level fine-grained guidance for improving synthesized results.

When retrieval-based guidance is used for feature normalization, it is difficult to exploit perceptual supervision for training, due to that the ground-truth image corresponding to retrieval-based guidance is missing.  
On the contrary, the ground-truth image of a semantic map can be naturally treated as a retrieval-based guidance, while the ground-truth image itself can also be used to facilitate perceptual supervision. 
However, ground-truth image is quite different from real retrieval-based guidance, and using it as guidance cannot make the learned model generate better synthesis results in the testing stage.
Instead, we introduce a data distortion mechanism on ground-truth images to mimic the quality of retrieval-based guidance.
%
%
During training, the distorted ground-truth images are also used as alternatives of retrieval-based guidance, making it feasible to leverage perceptual supervision for improving model training and visual quality.
Experiments on several challenging datasets show that our RESAIL performs favorably against state-of-the-arts.
The contributions of this work are summarized as:
\vspace{-0.6em}
\begin{itemize}
	\setlength{\itemsep}{0pt}
	\setlength{\parsep}{0pt}
	\setlength{\parskip}{0pt}
	\item A novel retrieval-based synthesis model is proposed by leveraging the retrieval-based guidance as pixel level fine-grained modulation, \ie, Retrieval-based Spatially Adaptive Normalization (RESAIL), for semantic image synthesis.
	\item During training, a data distortion mechanism on the ground-truth images is introduced to facilitate model training and improves visual quality of synthesized results.
	\item Extensive experiments show the effectiveness of our proposed method in synthesizing photo-realistic image from given semantic map.
\end{itemize}

\begin{figure*}
    \vspace{-0.4em}
	\centering
	\includegraphics[width=0.99 \linewidth]{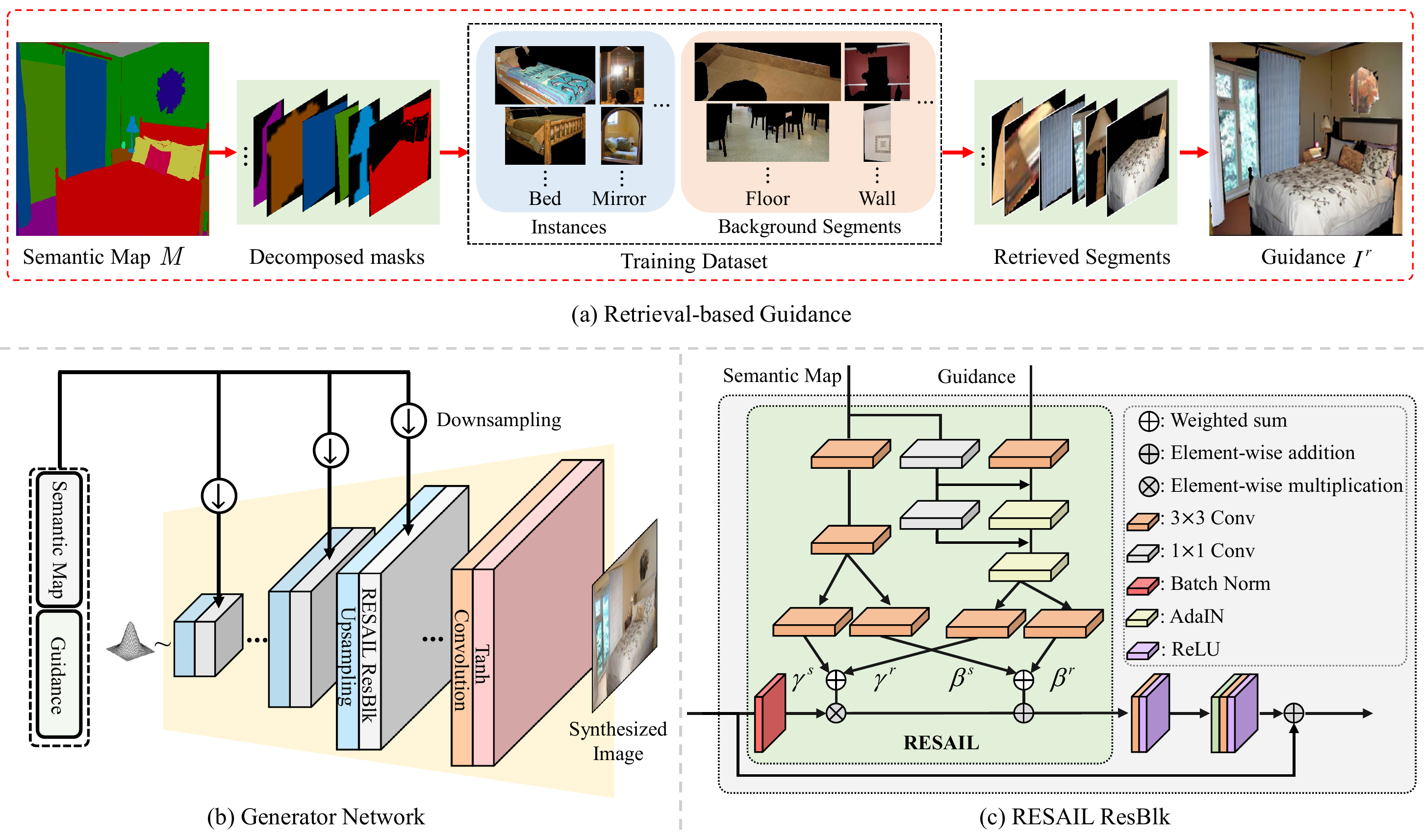}
	\vspace{-0.8em}
	\caption{Illustration of our method. (a) Given a semantic map $M$, we first retrieve a set of segments from the training dataset according to each semantic region of $M$ and composite them into the retrieval-based guidance $I^r$. It provides a pixel-level fine-grained guidance for the semantic image synthesis. (b) The architecture of our generator. It takes the semantic map and guidance as input, and consists of several RESAIL ResBlocks following upsample layers. (c) Detailed architecture of the RESAIL ResBlock used in (b). It learns the pixel level fine-grained modulation parameters from the semantic map and guidance for modulating the normalized activations.}
	\label{fig:training_architecture}
	\vspace{-1.5em}
\end{figure*}


\vspace{-0.4em}
\section{Related Work}
\vspace{-0.4em}
\subsection{Semantic Image Synthesis}
\vspace{-0.4em}

Many methods have been proposed to tackle semantic image synthesis. Here we focus on GAN-based methods, and also list other related methods~\cite{chen2017photographic, qi2018semi, li2019diverse}. 

Pix2pix~\cite{isola2017image} proposed a general framework for image-to-image translation, and Pix2pixHD~\cite{wang2018high} improved it for generating high-resolution images. In these methods, the semantic map is simply used as input to the network. SPADE~\cite{park2019semantic} exploited the semantic maps to predict transformation parameters for modulating the activations in normalization layers. Auxiliary guidance (\eg, style map~\cite{zhu2020sean} or 3D noise map~\cite{schonfeld2021you}) are incorporated with semantic map for diverse synthesis and easier controlling (details of normalization layer are surveyed in Sec.~\ref{sec:norm}). Instead of injecting semantic map into the network directly, CC-FPSE~\cite{liu2019learning} and SC-GAN~\cite{wang2021image} leveraged semantic map to predict the external parameters (convolution kernels~\cite{liu2019learning} or semantic vectors~\cite{wang2021image}), which are further used by another network to guide the image synthesis. 

Elaborate networks have also been explored in semantic image synthesis. SPADE~\cite{park2019semantic} employed a generator consisting of several residual blocks with upsampling layers and the PatchGAN discriminator. LGGAN~\cite{tang2020local} explored the local context information and introduce a local pathway in the generator for details synthesizing. CC-PFSE~\cite{liu2019learning} and SC-GAN~\cite{wang2021image} employed two generators for coarse and fine image synthesis. Besides generator, CC-FPSE~\cite{liu2019learning} proposed a feature-pyramid discriminator for semantically aligned image synthesis. SESAME~\cite{ntavelis2020sesame} and OSAIS~\cite{schonfeld2021you} improved the PatchGAN discriminator with a semantics-related mechanism. In addition, CollogeGAN~\cite{li2021collaging} used the StyleGAN~\cite{karras2019style} as the generator to improve visual quality and explored the local context with class-specific models. 

Among these methods, CC-FPSE and SC-GAN first synthesize a coarse image and use it to guide the fine image synthesis. While our method directly uses retrieval-based guidance to facilitate pixel level fine-grained modulation on activations.

\vspace{-0.5em}
\subsection{Conditional Normalization \label{sec:norm}}
\vspace{-0.5em}

Conditional normalization~\cite{dumoulin2016learned, huang2017arbitrary, park2019semantic, zhu2020sean} has been extensively studied in conditional image synthesis. 
Different from the earlier normalization techniques, conditional normalization layers require external data to learn the affine transformation parameters which are then used to modulate the normalized activations.
For example, Conditional Instance Normalization (CIN)~\cite{dumoulin2016learned} modified the $\gamma$ and $\beta$ parameters of Instance Normalization (IN) from length $C$ vectors to $N \times C$ matrices, and the external style $s$ is used to index the row of $\gamma$ and $\beta$. 
AdaIN \cite{huang2017arbitrary} learned a neural network that mapping the given style vectors to the $\gamma$ and $\beta$ parameters of IN. 
CIN and AdaIN perform uniformly across spatial coordinates, which may not be beneficial for the spatially-varying synthesis tasks, such as semantic image synthesis. 
Instead, SPADE~\cite{park2019semantic} proposed to learn a spatially-varying affine transformation in the semantic class level.
SEAN~\cite{zhu2020sean} extended the SPADE with a style map which is composed of the style vectors for each region, and learned the transformation parameters from both semantic map and the style map in the region level.
%
%
OASIS~\cite{schonfeld2021you} introduced a 3D noise concatenating with the semantic map to perform the spatially-variant normalization, but the 3D noise provides limited semantic information for the synthesis.
CLADE~\cite{tan2021efficient} learned a parameter bank for each semantic class, which is used to generate the parameters for modulation, but still limited to coarse-level guidance.

In contrast, our RESAIL module takes the retrieved results to introduce pixel level fine-grained guidance for semantic image synthesis.

\vspace{-0.4em}
\subsection{Retrieval-based Image Synthesis}
\vspace{-0.2em}

In the early studies, many retrieval-based methods \cite{hays2007scene,lalonde2007photo,chen2009sketch2photo,johnson2010cg2real,isola2013scene} have been proposed for conditional image synthesis. 
For example, Hays \etal \cite{hays2007scene} used a collection of images as retrieval database for image completion. In testing stage, similar images are retrieved via the descriptor matching and used to complete the missing regions. 
Lalonde \etal \cite{lalonde2007photo} retrieved object segments from a large image database and then interactively composited them into an image. 
Chen \etal \cite{chen2009sketch2photo} developed a system that retrieved and synthesized an image from a freehand sketch with associated text labels.
Isola and Liu \cite{isola2013scene} presented an analysis-by-synthesis method that retrieved segments according to the given query image and combined these segments to form a “scene collage” that explains the query.
%
%
Recently, SIMS~\cite{qi2018semi} leveraged deep network for improving the quality of synthesized results.
However, it simply takes the retrieved image as network input, failing in exploiting progress in conditional normalization. In contrast, our method uses retrieval-based guidance for spatially adaptive normalization, which is beneficial for synthesizing photo-realistic images.

\begin{figure}
	\centering
	\includegraphics[width=0.99 \linewidth]{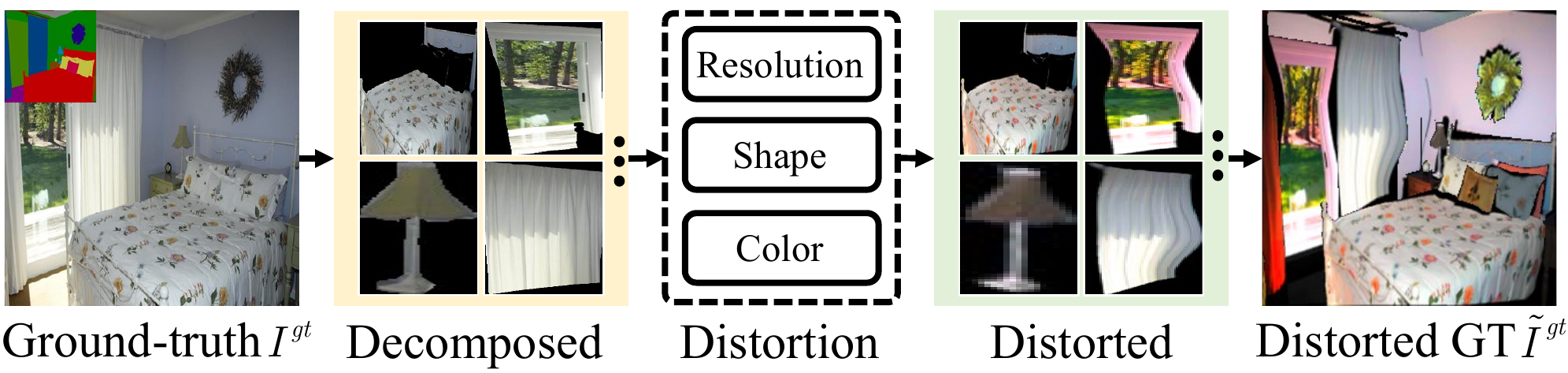}
	\vspace{-0.2em}
	\caption{Illustration of data distortion on ground-truth image $I^{gt}$. Specifically, $I^{gt}$ is first decomposed into several segments based on semantic map. Then each segment is distorted separately by modifying shape, color, and resolution. Finally, distorted segments are composited into a distorted ground-truth image $\tilde{I}^{gt}$.}
	\label{fig:data_distort}
	\vspace{-1.7em}
\end{figure}

\vspace{-0.3em}
\section{Proposed Method}
\vspace{-0.2em}



%
Given a semantic map $M \in \{0, 1\}^{H \times W \times C}$, semantic image synthesis aims to generate the corresponding images $\hat{I} \in \mathbb{R}^{H \times W \times 3}$. Here $H$, $W$, and $C$ denote the height, width, and number of categories in semantic map, respectively.
%
%
In this section,
we first present a retrieval paradigm 
to produce a retrieval-based guidance $I^r$ (Sec.~\ref{sec:retrieval}). 
We also introduce the distorted ground-truth as the alternative of retrieval-based guidance, and introduce the perceptual supervision to facilitate model training, (Sec.~\ref{sec:distortion}). 
With the guidance, we propose a Retrieval-based Spatially Adaptive Normalization (RESAIL) to perform pixel level fine-grained modulation on activations (Sec.~\ref{sec:resail}).
Finally, we introduce several loss terms for training the model to generate the photo-realistic images (Sec.~\ref{sec:losses}).

\vspace{-0.3em}
\subsection{Retrieval-based Guidance \label{sec:retrieval}} 
\vspace{-0.2em}

Given the semantic map $M$, we first present a retrieval paradigm to obtain the retrieval-based guidance from the training dataset which contains pixel level fine-grained information.
%
%
%
As shown in Fig.~\ref{fig:training_architecture}(a), the semantic map $M$ can be decomposed into several object segment masks $M = \{(M^s_i, y^c_i)\} $, where $M^s_i$ denotes the cropped binary segment mask of one object and $y^c_i$ is the corresponding category. 
Similarly, a training image can also be decomposed into segment images according the semantic map. 
We define these segments as the retrieval unit. 
In training or testing stage,  the retrieval-based guidance is obtained by,
\vspace{-0.3em}
\begin{equation}
	I^r = \Theta \left(\{\Gamma(\mathcal{D}^{tr}, M^s_i, y^c_i) ~ | ~ (M^s_i, y^c_i) \in M \} \right)
    \label{eq:retrive}
    \vspace{-0.3em}
\end{equation}
where $\Gamma(\mathcal{D}^{tr}, M^s_i, y^c_i)$ denotes the retrieval function defined on training dataset $\mathcal{D}^{tr}$. 
It finds a segment image with category $y^c_i$ and the most similar shape with $M^s_i$.
When there is no matching segment image in training dataset, we replace it with a black image.
$\Theta(\cdot)$ function recomposes the retrieved segments to form the guidance.
%
%
Note that, in the training stage, we ignore the original segment images corresponding to $M$ and retrieve the other most compatible segment images based on the geometric consistency score~\cite{wang2020constrained}.
More details are provided in the \emph{Suppl}.

\vspace{-0.3em}
\subsection{Distorted Ground-truth as Guidance \label{sec:distortion}}
\vspace{-0.2em}
The retrieval-based guidance image $I^r$ lacks of paired ground-truth, making it impossible to exploit perceptual supervision during training. 
Intuitively, the ground-truth image can be used as both the guidance and the ground-truth, resulting a paired training data. %
However, ground-truth image is quite different from real retrieval-based guidance (\eg, color, shape and resolution distortion usually are inevitable in retrieval-based guidance, see Fig.~\ref{fig:training_architecture}(a)).
Thus, directly using ground-truth as guidance in training benefits little to learn generator that works well for retrieval-based guidance.
Instead, we introduce a data distortion mechanism on ground-truth images to mimic the quality of retrieval-based guidance.
As illustrated in Fig.~\ref{fig:data_distort}, the ground-truth is first decomposed into a set of separate segments. 
Then these segments are distorted by changing shape, color and resolution, respectively. 
Finally, the distorted segment images are recomposed into the distorted ground-truth $\tilde{I}^{gt}$, which can be utilized as alternative of retrieval-based guidance.
Due to that the distorted ground-truth has the real paired image (\ie, original ground-truth), we can introduce perceptual supervision on synthesis results to facilitate model training and improve visual quality.

\begin{figure*}
\vspace{-1.2em}
    \label{fig:comparsions}
	\centering
	\includegraphics[width=0.99 \linewidth]{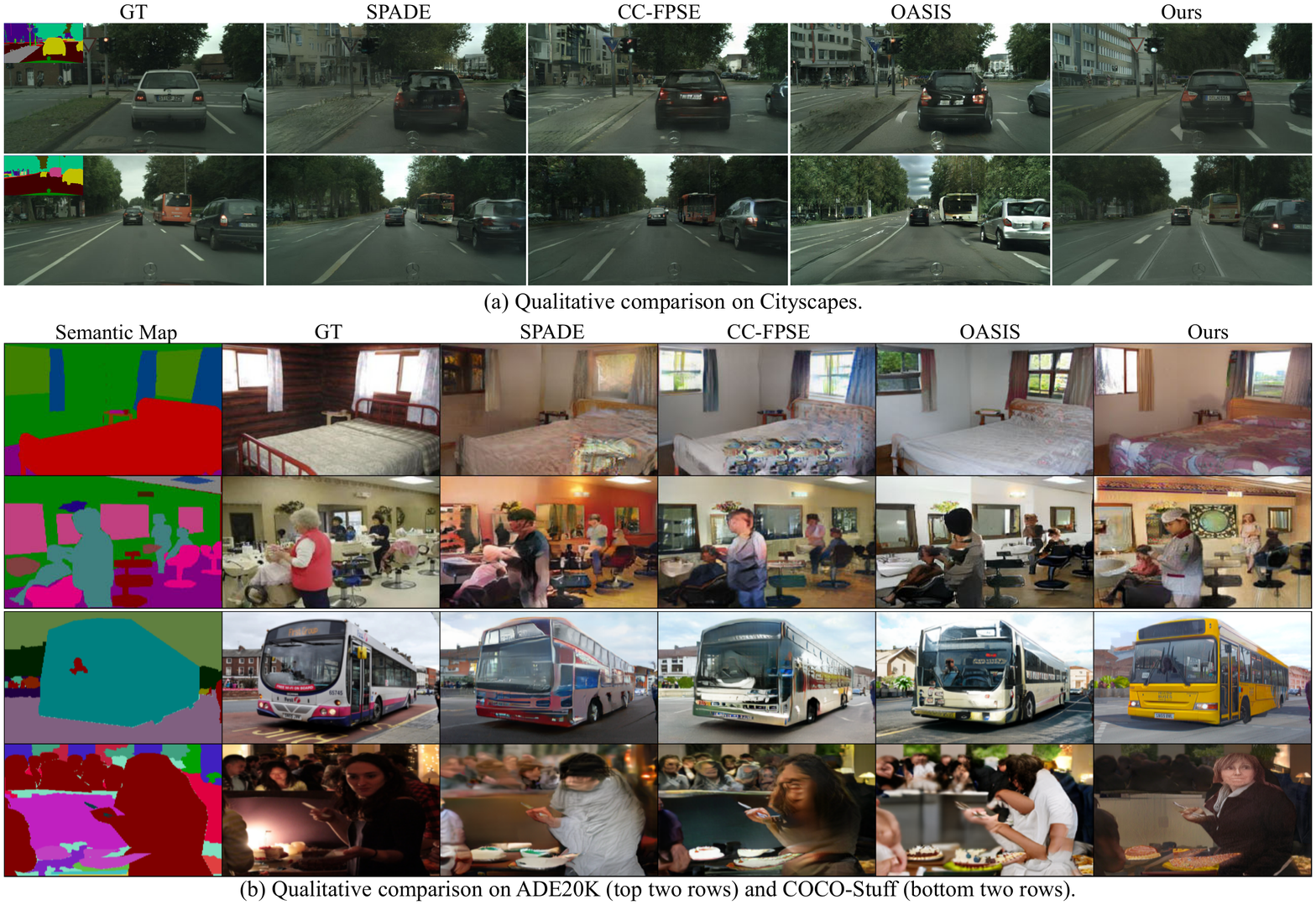}
	\vspace{-0.8em}
	\caption{Qualitative comparison of our method with the competing methods on the (a) Cityscapes, (b) ADE20K and COCO-Stuff datasets. Our model generates images with better perceptual quality and finer details.}
	\label{fig:comparsions}
	\vspace{-1.2em}
\end{figure*}

\vspace{-0.3em}
\subsection{Network Architecture \label{sec:resail}}
\vspace{-0.2em}

\noindent \textbf{Retrieval-based Spatially Adaptive Normalization.} 
With the guidance $I^r$ (or $\tilde{I}^{gt}$) and semantic map $M$, we propose a REtrieval-based Spatially AdaptIve normaLization (RESAIL) to perform pixel level fine-grained modulation on feature activations. 
Specifically, we adopt the conditional normalization architecture with spatially adaptive modulation.
As the guidance image contains pixel level information about the object class, we first use it to learn the \emph{fine-grained modulation parameters} (\ie, $\gamma^r$ for scale and $\beta^r$ for bias) by a four-layer convolutional network. 
Due to there are some semantic regions missing in the retrieval-based guidance image (no matching segment images or shape gaps), 3$\times$3 kernel is used in the convolutional layer to complete the information in the missing region. 
Besides, we use the AdaIN incorporated with the semantic map in the intermediate two layers to further enrich the semantic information of the missing area.
The detailed structure is shown in Fig.~\ref{fig:training_architecture}(c).
Analogous to~\cite{park2019semantic, zhu2020sean}, we also learn the \emph{coarse modulation parameters} (\ie, $\gamma^s$ and $\beta^s$) from the semantic map.
Two sets of parameters are weighted summed to get the final pixel level fine-grained modulation parameters,
\vspace{-0.4em}
\begin{equation}
\begin{split}
    \gamma &= \alpha_{\gamma}  \gamma^s + \left(1 - \alpha_{\gamma} \right) \gamma^ r, \\
	\beta  &= \alpha_{\beta}  \beta^s + \left(1 - \alpha_{\beta} \right) \beta^r,
\end{split}
\end{equation}
where $\alpha_\gamma$ and $\alpha_\beta$ are learnable weight parameters, and the input activations are finally modulated by, 
\vspace{-0.4em}
\begin{equation}
\textit{RESAIL}(\mathbf{h}, M, I^r) = \gamma_{c, y, x}  \frac {\mathbf{h}_{n, c, y, x} - \mathbf{\mu}_c} { \mathbf{\sigma}_c} + \beta_{c, y, x}\,,
\label{eq: activation}
\end{equation}
where $ \mathbf{h} $ denotes the input activations with a batch of $N$ samples, $\mu$ and $ \sigma $ denote the mean and standard deviation of the activations. 
$ \left( n \in N, c \in C , y \in H, x \in W \right)$ sites the modulated activations value. More details about the RESAIL module are provided in the \emph{Suppl}.

\noindent \textbf{Generator.}~Fig.~\ref{fig:training_architecture}(b) illustrates the architecture of our generator $G$, which is built on the generator of SPADE \cite{park2019semantic}. 
Analogous to \cite{park2019semantic}, we employ a generator consisting of several RESAIL residual blocks (RESAIL ResBlk) with upsampling layers. The semantic map $M$ and guidance ($I^r$ or $ \tilde{I}^{gt}$) are resized and fed to each RESAIL module to guide the image synthesis,
\begin{equation}
    \vspace{-0.2em}
    \begin{split}
        \hat{I} = G(M, I^r),  \quad \hat{I}^{gt} = G(M, \tilde{I}^{gt}).
    \end{split}
\vspace{-0.2em}
\end{equation}
\vspace{-2.0em}
\subsection{Loss Functions \label{sec:losses}}
\vspace{-0.5em}

As discussed above, we first introduce the perceptual loss $\mathcal{L}_{vgg}$ \cite{wang2018high} and feature matching loss $\mathcal{L}_{FM}$ \cite{wang2018high} between $I^{gt}$ and the synthesized image $\hat{I}^{gt}$ to facilitate the model training. 
To encourage the generator to synthesize photo-realistic images, we also introduce the adversarial loss \cite{ntavelis2020sesame} on synthesized images (both $\hat{I}$ and $\hat{I}^{gt}$).
Besides, to emphasize the synthesis of each semantic region, we incorporate a segmentation loss with the model training. 
Specifically, we introduce a pretrained segmentation network $ S $ to classify the category of each entry on the generated image,
\vspace{-0.2em}
\begin{equation}
	\mathcal{L}_{cls} = - \mathbb{E}_{M}  \left[ \sum  \limits_c \alpha_c \sum \limits_{i, j} M_{i,j,c} \log{ S(\hat{I})_{i,j,c} } \right] ,
	\label{eq:cls}
	\vspace{-0.2em}
\end{equation}
where $ \alpha_c $ denotes the class balancing weight~\cite{schonfeld2021you}, and $S$ is pretrained on the training dataset. 
$\mathcal{L}_{cls}$ is introduced on both $\hat{I}$ and $\hat{I}^{gt}$. 
Finally, we combine all the above losses to give the overall learning objective,
\vspace{-0.2em}
\begin{equation}
	\mathcal{L} = \lambda_{vgg}\mathcal{L}_{vgg} + \lambda_{fm}\mathcal{L}_{fm} + \lambda_{adv}\mathcal{L}_{adv} + \lambda_{cls}\mathcal{L}_{cls},
	\label{eq:loss}
	\vspace{-0.2em}
\end{equation}
where $\lambda_*$ denotes tradeoff parameters for different losses.

\begin{figure*}
\vspace{-0.8em}
	\centering
	\includegraphics[width=0.99 \linewidth]{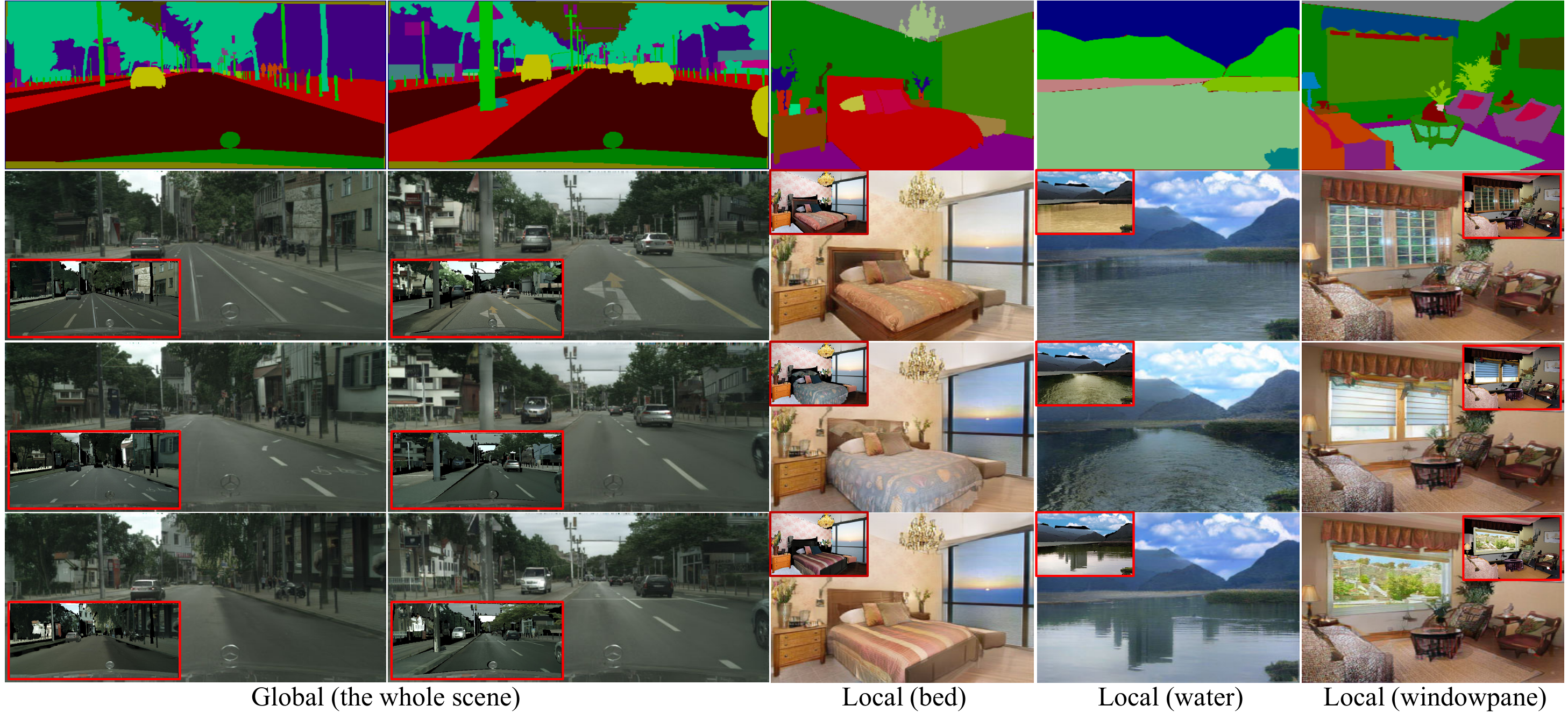}
	\vspace{-0.8em}
	\caption{Multi-modal synthesis capability of our method. Each column represents the synthesized results with the given semantic map (top row). During testing, we retrieve a set of different guidance images, resulting diverse synthesized images (\ie, left 2 columns). We can also fix most semantic regions and change the retrieved segments of certain objects to achieve local editing results (\ie, right 3 columns). The retrieval-based guidance images used for the image synthesis are given in the red rectangle.}
	\label{fig:multi-results}
	\vspace{-1em}
\end{figure*}

\begin{figure*}
	\centering
	\includegraphics[width=0.99 \linewidth]{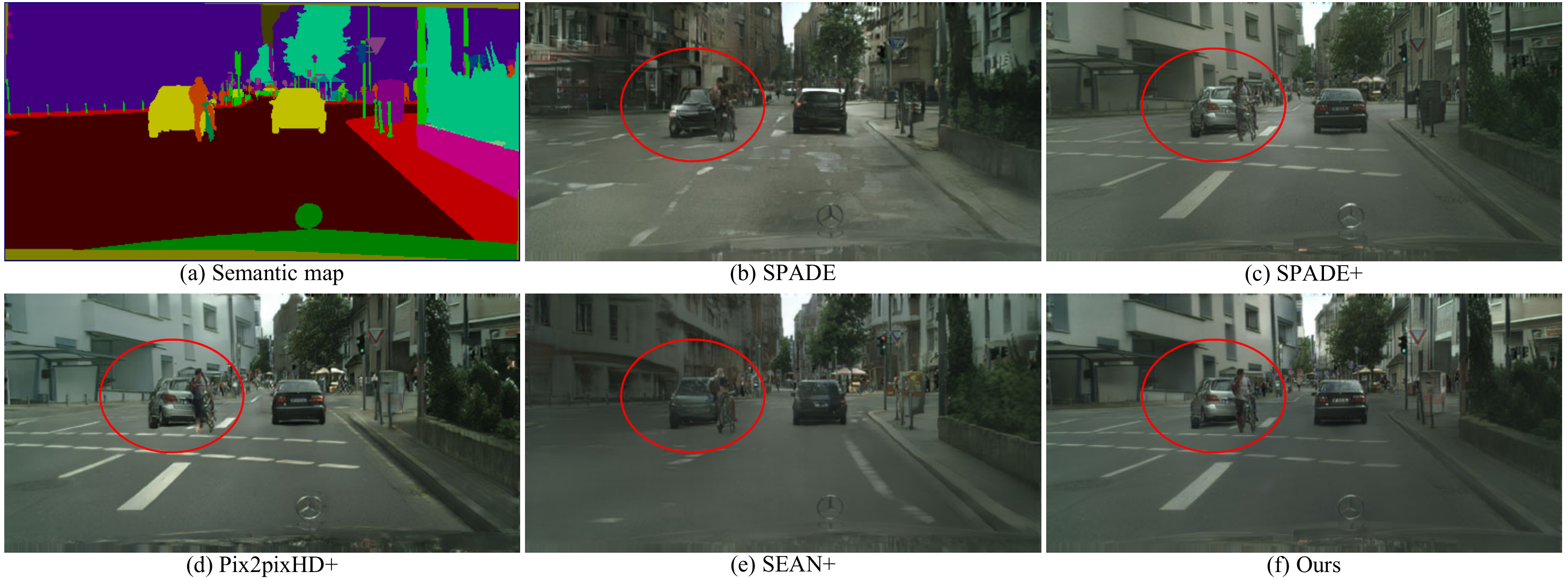}
	\vspace{-0.9em}
	\caption{Ablation study on the RESAIL module. Model+ denotes introducing the retrieval-based guidance to the model (see Sec.~\ref{sec:ablation} and the \emph{Suppl} for more details). With the proposed RESAIL module and the retrieval-based guidance, our method produces more photo-realistic details (red circle). Zoom for a better view.}
	\label{fig:structure_aba}
	\vspace{-1.8em}
\end{figure*}

\vspace{-0.3em}
\section{Experiments}
\vspace{-0.2em}


\subsection{Experimental Settings}
\vspace{-0.4em}

\noindent \textbf{Datasets.}~We evaluate our model on four common used datasets, Cityscapes~\cite{cordts2016cityscapes}, ADE20K~\cite{zhou2017scene}, ADE20K-outdoor and COCO-Stuff~\cite{caesar2018coco}.
The training set of Cityscapes consists of 3,000 images, including 35 semantic categories, while the validation set consists of 500 images. 
The ADE20K dataset contains over 20,000 images for training and 2,000 images for validation with 150 semantic classes in total.
The ADE20K-outdoor dataset is a subset of ADE20K only containing outdoor scenes.
COCO-Stuff consists of 118,000 training images and 5,000 validation images.

\noindent\textbf{Evaluation Metric.}~Pixel ACcuracy (AC) and mean Intersection-Over-Union (mIOU) are adopted, which measure the agreement between synthesized image and given input~\cite{chen2017photographic, park2019semantic, ntavelis2020sesame}. 
They both require a pretrained segmentation model to compute segmentation accuracy  \cite{yu2017dilated, xiao2018unified, chen2017deeplab}.
We also utilize Frechet Inception Distance (FID) \cite{heusel2017gans} to evaluate the quality of synthesized images. 

\noindent \textbf{Implementation Details.}~We train our model on four Tesla v100 GPUs and adopt ADAM optimizer with $ \beta_1 = 0$ and $ \beta_2 = 0.999$ where the learning rates are set to 0.0001 for generator and 0.0004 for discriminator.
Additionally, we apply the spectral normalization~\cite{miyato2018spectral} to each layer in both generator and discriminator, and use synchronized BatchNorm \cite{zhang2018context} in RESAIL blocks.

\begin{table*}[t]
	{\small
		\begin{center}
		\vspace{-0.8em}
		\caption{Quantitative comparison on ADE20K~\cite{zhou2017scene}, ADE20K-outdoor,  Cityscapes~\cite{cordts2016cityscapes} and COCO-Stuff~\cite{caesar2018coco}. For AC and mIOU, higher is better, and for FID, lower is better. Our method achieves very competitive results on the four datasets.}
		\vspace{-0.8em}
		\label{tab:results}
		\resizebox{0.99\linewidth}{!}{
			\aboverulesep=0ex
			\belowrulesep=0ex
			\renewcommand{\arraystretch}{1.2}
			\begin{tabular}{lcccccccccccc}
				\toprule
				\multirow{2}[3]{*}{Method} &
				\multicolumn{3}{c}{ADE20K} &
				\multicolumn{3}{c}{ADE20K-outdoor} &
				\multicolumn{3}{c}{Cityscapes} &
				\multicolumn{3}{c}{COCO-Stuff} \\
				\cmidrule(llr){2-4}
				\cmidrule(llr){5-7}
				\cmidrule(llr){8-10}
				\cmidrule(llr){11-13}
				& FID ($ \downarrow $) & AC ($ \uparrow $) & mIOU ($ \uparrow $) & FID ($ \downarrow $) & AC ($ \uparrow $) & mIOU ($ \uparrow $) & FID ($ \downarrow $) & AC ($ \uparrow $) & mIOU ($ \uparrow $) & FID ($ \downarrow $) & AC ($ \uparrow $) & mIOU ($ \uparrow $) \\
				\midrule
				CRN~\cite{chen2017photographic} &  73.3 & 68.8 & 22.4 & 99.0 & 68.6 & 16.5 & 104.7 & 77.1 & 52.4 & 70.4 & 40.4 & 23.7 \\
				
				Pix2pixHD~\cite{wang2018high}  & 81.8 & 69.2 & 20.3 & 97.8 & 71.6 & 17.4 & 95.0 & 81.4 & 58.3 & 111.5 & 45.7 & 14.6 \\
				
				SIMS~\cite{qi2018semi} & n/a & n/a & n/a & 67.7 & 74.7 & 13.1 & 49.7 & 75.5 & 47.2 & n/a & n/a & n/a \\
				
				SPADE~\cite{park2019semantic}  & 33.9 & 79.9 & 38.5 & 63.3 & 82.9 & 30.8 & 71.8 & 81.9 & 62.3  & 22.6 & 67.9 & 37.4 \\
				
				CC-FPSE~\cite{liu2019learning} & 31.7 & 82.9 & 43.7 & n/a & n/a & n/a & 54.3 & 82.3 & 65.5 & 19.2 & 70.7 & 41.6 \\

				SESAME ~\cite{ntavelis2020sesame} & 31.9 & \textbf{85.5} & 49.0 & n/a & n/a & n/a & 54.2 & 82.5 & 66.0 & n/a & n/a & n/a \\
				
				SC-GAN~\cite{wang2021image}  & 29.3 & 83.8 & 45.2 & n/a & n/a & n/a & 49.5 & 82.5 & 66.9 & 18.1 & 72.0 & 42.0 \\
				
				OASIS~\cite{schonfeld2021you} & \textbf{28.3} & n/a & 48.8 & \textbf{48.6} & n/a & 40.4 & 47.7 & n/a & 69.3 & \textbf{17.0} & n/a & 44.1 \\
				
				Ours & 30.2 & 84.8 & \textbf{49.3} & \textbf{48.6} & \textbf{86.5} & \textbf{41.1} & \textbf{45.5} & \textbf{83.2} & \textbf{69.7} & 18.3 & \textbf{73.1} & \textbf{44.7} \\
				
				\bottomrule
				\vspace{0.1pt}
			\end{tabular}
		}
		\end{center}
	}
	\vspace{-2.4em}
\end{table*}

\vspace{-0.4em}
\subsection{Qualitative Results}
\vspace{-0.4em}

\begin{figure*}
	\centering
	\vspace{-0.7em}
	\includegraphics[width=0.99 \linewidth]{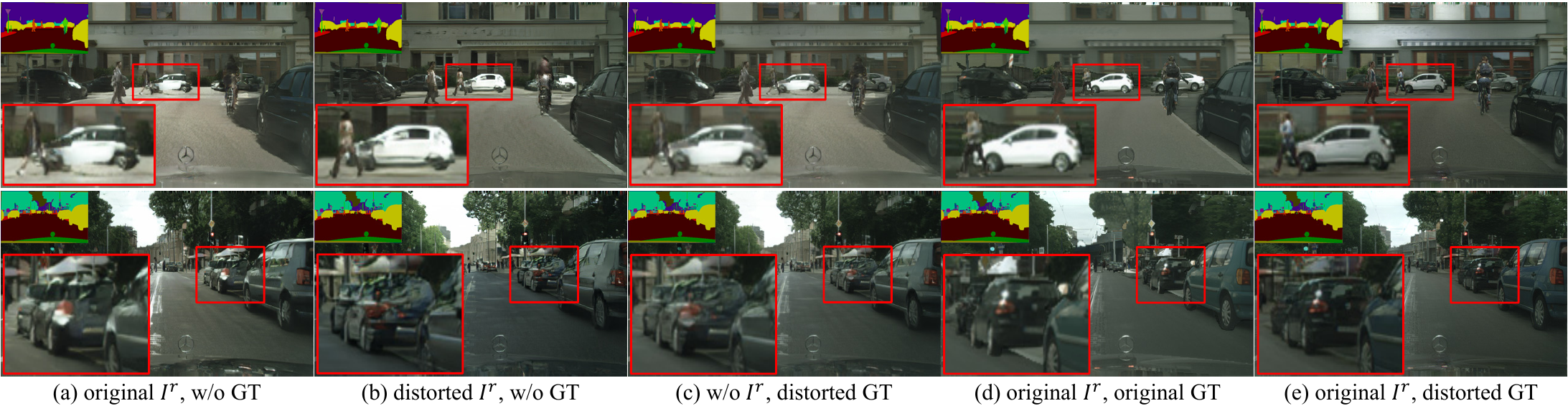}
	\vspace{-0.8em}
	\caption{Ablation study on the data distortion method. (a)(b) When only retrieval-based guidance $I^r$ is used in training, generator fails to synthesize certain objects details marked in red rectangle; (c) Synthesized images also suffer from poor details only with the guidance of \emph{distorted GT}. (d) Using both $I^r$ and \emph{original GT} as guidance, inconsistent edge and illumination can still be observed. 
	(e) Using both $I^r$ and \emph{distorted GT} $\tilde{I}^{gt}$ as guidance, our model synthesizes photo-realistic image with fine details. Please zoom for a better view.}
	\label{fig:data_aba}
	\vspace{-1.7em}
\end{figure*}

%


We first qualitatively compare our model with the state-of-the-art methods~\cite{park2019semantic, schonfeld2021you, liu2019learning} on the Cityscapes, ADE20K and COCO-Stuff datasets, and the results are illustrated in Fig.~\ref{fig:comparsions}. 
For SPADE~\cite{park2019semantic} and CC-FPSE~\cite{liu2019learning}, degenerated synthesis results on some objects can be observed, such as car and bed. 
Although OASIS~\cite{schonfeld2021you} introduces semantic discriminator to improve the visual quality of synthesized image, it is still limited in avoiding unrealistic details and obvious artifacts.
In contrast, 
%
benefited from the retrieval-based guidance, our model generates more photo-realistic images with finer details such as edges, textures, color, and less artifacts.

Moreover, we retrieve the segment image for each semantic region separately, which allows us to edit the synthesis images either globally or locally.
As shown in Fig.~\ref{fig:multi-results}, given the same semantic map, we can 
achieve globally diverse synthesis results by changing all the retrieved segments of the whole image (left two columns). 
Moreover, We can also fix most semantic regions and change the retrieved segments of the remaining objects to edit the results locally (right three columns).
%
%
More qualitative results are provided in the \emph{Suppl}.
%


\vspace{-0.4em}
\subsection{Quantitative Results}
\vspace{-0.4em}
We further quantitatively compare with the competing methods~\cite{chen2017photographic, wang2018high, qi2018semi, park2019semantic, liu2019learning, ntavelis2020sesame, wang2021image, schonfeld2021you} on four datasets, and Table~\ref{tab:results} lists the results.
From the table, our method performs favorably against the competing methods on  Cityscapes~\cite{cordts2016cityscapes} and ADE20K-outdoor datasets, and also is very competitive on ADE20K~\cite{zhou2017scene} and COCO-Stuff~\cite{caesar2018coco} datasets,  demonstrating the effectiveness of our method. 
Note that, SIMS~\cite{qi2018semi} also uses a retrieved image to guide semantic image synthesis but is inferior to our method, partially due to that it is more effective to use retrieval-based guidance for spatially adaptive normalization other than use it as network input.

\begin{table}[t]

\begin{center}
	\caption{User study on Cityscapes. The numbers indicate the percentage of volunteers who favor the results of our method over those of the competing methods or even the ground-truth. }
	\vspace{-0.8em}
	\label{tab:user_study}
	\resizebox{1\linewidth}{!}{
		\renewcommand{\arraystretch}{1.2}
		
		\begin{tabular}{cccc}
			
			\toprule
			Ours vs. SPADE & Ours vs. CC-FPSE  & Ours vs. OASIS & Ours vs. GT  \\
			\midrule
			87.8 & 80.2 & 85.4 & 16.8 \\
			\bottomrule
			
		\end{tabular}
	}
\end{center}
\vspace{-3.1em}
\end{table}

\noindent \textbf{User Study.}  Following the previous works \cite{park2019semantic, ntavelis2020sesame, wang2021image}, we conduct user study on Cityscapes dataset.
%
Participants have been informed their identities will not be recorded.
%
Each volunteer is given a semantic map and two corresponding images containing one by our method and another one by a randomly selected competing method (\ie, SPADE~\cite{park2019semantic}, CC-FPSE~\cite{liu2019learning}, OASIS~\cite{schonfeld2021you} or even the ground-truth image), and is asked to vote for the image with better visual quality. 
The orders of the two images are random to avoid the effect caused by potential bias. 
There are totally 2,000 questions for 200 volunteers, and Table~\ref{tab:user_study} lists the results.
Volunteers strongly favor (more than 80\%) our results in contrast to the competing methods.
%
In comparison with the ground-truth images, our results still have a chance of about 17\% to be recognized as the better one, further indicating our method is able to generate photo-realistic images.

\begin{table}[t]
\small{
\begin{center}
    \vspace{-0.5em}
	\caption{Ablation study on RESAIL module. Model+ denotes using the retrieval-based guidance as input to the given module. With the proposed RESAIL module and the retrieval-based guidance, our method achieves better quantitative performance.}
	\vspace{-1em}
	 \resizebox{1.0\linewidth}{!}{
	
	\label{tab:structure_ablations_results}
	    \setlength{\tabcolsep}{2.5mm}
	    \begin{tabular}{ccccc}
			\toprule
			Variants & Guidance Inject & FID$\left(\downarrow\right)$ & mIOU$\left(\uparrow\right)$ & AC$\left(\uparrow\right)$  \\
			\midrule
			SPADE & w/o & 58.7 & 62.2 & 81.9 \\
			
			Pix2pixHD+ & Conv Layer & 47.8 & 66.7 & 81.9 \\
			SPADE+ & SPADE Module & 53.4 & 68.6 & 82.8 \\
			SEAN+ & SEAN Module & 66.6 & 69.4 & 82.1 \\
		
			Ours & RESAIL Module & \textbf{45.5} & \textbf{69.7} & \textbf{83.2} \\
							
			\bottomrule
			\vspace{0.1pt}
		\end{tabular}
    	}
    	
\end{center}
}
\vspace{-3em}
\end{table}

\begin{table}[t]
\small{
\begin{center}
    \caption{Effect of data distortion mechanism on ground-truth guidance. Among all variants, using both $I^r$ and distorted ground-truth $\tilde{I}^{gt}$ as guidance achieves better performance.}
    \vspace{-1em}
	\label{tab:data_ablations_results}
	\begin{tabular}{cc|ccc}
		\toprule
		 $I^r$ &  Ground-truth & FID $(\downarrow)$ & mIOU $(\uparrow)$ & AC $(\uparrow)$  \\
		\midrule
	
		original & w/o  & 47.7 & 66.3 & 82.5 \\
		distorted & w/o  & 48.8 & 65.3 & 82.6 \\
		w/o & distorted  & 49.0 & 64.9 & 82.1 \\
		original & original & 52.8 & 64.0 & 81.2 \\
		original & distorted & \textbf{45.5} & \textbf{69.7} & \textbf{83.2} \\

		\bottomrule

	\end{tabular}
	
\end{center}
\vspace{-3em}
}
\end{table}

\vspace{-0.3em}
\subsection{Ablation Studies \label{sec:ablation}}
\vspace{-0.2em}

We conduct ablation studies on Cityscapes to assess the effect of RESAIL module and data distortion mechanism.


\noindent \textbf{Effectiveness of RESAIL Module.} 
To demonstrate the effectiveness of our RESAIL module,
we compare our method with 4 variants which vary on whether the retrieval-based guidance used and how to use it: 
(i) \textit{SPADE} denotes the original SPADE module without exploiting the guidance. 
(ii) \textit{Pix2pixHD+} denotes concatenating the guidance into the conv layer of pix2pixHD model. 
(iii) \textit{SPADE+} denotes using the guidance as input to the SPADE module. 
(iv) \textit{SEAN+} denotes using the guidance as input to the SEAN module. 
(v) \textit{Ours} denotes using the guidance as input to the RESAIL module. 
More details about the architecture of each variant can be found in the \emph{Suppl}.  
For a fair comparison, we use the same backbone for all variants and only change the normalization layer. Thus for \textit{Pix2pixHD+}, we use the decoder part as the generator.

Table~\ref{tab:structure_ablations_results} lists the quantitative comparison among the variants. 
From the table, directly incorporating the guidance into the SPADE or conv layer improves the performance, indicating that the retrieval-based guidance is beneficial to image synthesis.
As for SEAN, regarding the style map is heavy in GPU memory-consuming, we reduce the dimension of style vector to 128 to conduct the experiments, which may cause potential performance degradation but not affect the fair comparison. 
With the RESAIL module, our method achieves the best performance, clearly demonstrating the effectiveness of our RESAIL module. 
%
%
As shown in Fig.~\ref{fig:structure_aba}, without pixel level guidance information, \textit{SPADE} and \textit{SEAN+} generate blurry details. 
In compared to \textit{Pix2pixHD+}, our RESAIL generates more photo-realistic results with finer details and consistent illumination. 
The result shows that spatially adaptive normalization is a more effective way to use retrieval-based  guidance than simply concatenating it with feature of conv layer.

\noindent \textbf{Effectiveness of Distorted Ground-truth.} 
We also conduct the ablation study to assess the effect of data distortion mechanism on ground-truth (GT) images.
Specifically, we consider five variants. 
(i) Only the retrieval-based guidance $I^r$ is used as guidance during training. 
(ii) Only the distorted $I^r$ is used as guidance during training. 
(iii) Only the distorted GT is used as guidance during training. 
(iv) Both $I^r$ and the original GT can be used as guidance during training. 
(v) Ours: both $I^r$ and distorted GT $\tilde{I}^{gt}$ can be used as guidance during training.
%
%

Table~\ref{tab:data_ablations_results} lists the quantitative results on Cityscapes. 
From the table, performing data distortion on retrieval-based guidance brings little gain or even adverse effect on semantic image synthesis. 
This is because the retrieval-based guidance is already distorted and further distorting it may make it more unrealistic and is not beneficial to synthesis performance. 
Also, using the original GT as guidance cannot improve the quality of generated images, because there exists obvious gap between original GT and retrieval-based guidance. 
With the data distortion on the ground-truth, we can reduce the gap between them and thus benefits the model training.
Fig.~\ref{fig:data_aba} shows the qualitative results. 
One can see that, using both retrieval-based guidance and distorted ground-truth as guidance during training, our method produces more photo-realistic details and consistent color.

Additional ablation study on the segmentation loss is provided in the \emph{Suppl}, please check it for more details.

\vspace{-0.6em}
\section{Discussion}
\vspace{-0.5em}
In this paper, we proposed a novel feature normalization method, termed as REtrieval-based Spatially AdaptIve normaLization (RESAIL). With the retrieval-based guidance and distorted ground-truth, the model can be trained with perceptual supervision, and produces diverse and photo-realistic synthesized images. Experimental results demonstrate that our method performs favorably against the state-of-the-art methods on several challenging datasets both qualitatively and quantitatively.


\noindent \textbf{Impact.} 
This work presents a RESAIL module for semantic image synthesis. Although we have not conducted the experiments on human face synthesis tasks,  it has the potential for being used to face synthesis and editing. From this viewpoint, our work may be improperly used for deepfake techniques which trigger potential negative social impacts.

\noindent \textbf{Limitation.}~Albeit our method synthesizes photo-realistic images and outperforms existing methods, the inference speed is still a limitation. Retrieving operation in our method is time consuming, which makes it unable to perform realtime inference. In the future, we will explore feasible method to accelerate or avoid the retrieving process.

\vspace{-0.5em}
	\section*{Acknowledgement}
	\vspace{-0.5em}
	
	This work was supported in part by National Key R\&D Program of China under Grant No. 2020AAA0104500, and by the National Natural Science Foundation of China (NSFC) under Grant No.s U19A2073 and 62006064.




\newpage
{\small
\bibliographystyle{ieee_fullname}
\bibliography{egbib}
}
\newpage


 
\twocolumn[
\begin{center}
	{\LARGE \textbf{Supplemental Materials}}
	\vspace{24pt}
\end{center}]
\setcounter{section}{0}
\setcounter{table}{0}
\setcounter{figure}{0}
\setcounter{equation}{0}
\renewcommand\thesection{\Alph{section}}
\renewcommand\thefigure{\Alph{figure}}
\renewcommand\thetable{\Alph{table}}
\renewcommand\theequation{\alph{equation}}
\renewcommand\thesubsection{\thesection.\arabic{subsection}}

\newcommand*{\affaddr}[1]{#1} 
\newcommand*{\affmark}[1][*]{\textsuperscript{#1}}
\newcommand*{\email}[1]{\texttt{#1}}




\section{Additional Implementation Details \label{sec:imp}}

\begin{figure*}
	\centering
	\includegraphics[width=0.99 \linewidth]{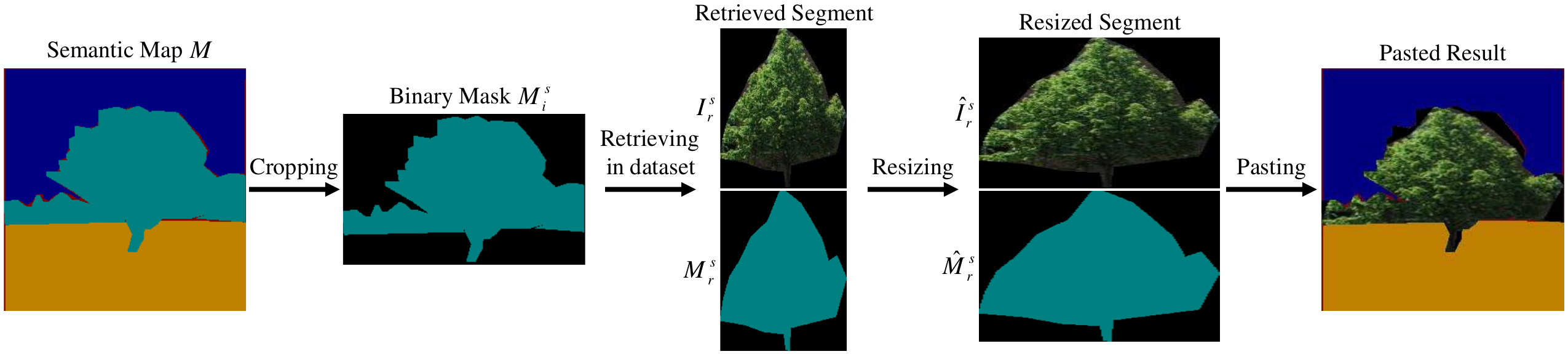}
	\caption{Process to paste a retrieved segment into the semantic map. We here take ``Tree" labeled in cyan as an example.}
	\label{fig:composition}
	\vspace{-0.8em}
\end{figure*}

\subsection{Retrieval-based Guidance Image}
Given a semantic map $M$, we use it to retrieve and composite a guidance image $I^r$ for image synthesis.

\noindent \textbf{Preprocess of Dataset.} 
The training dataset $\mathcal{D}^{tr}$ is firstly used to create a retrieval database consisting of a set of segments. 
Specifically, for each image $I \in \mathcal{D}^{tr}$ and its corresponding semantic map $M$, we use the available instance-level annotation to decompose $I$ and $M$ as a number of segments,
\begin{equation}
    I, M=\left\{\left(M_i^s, y_i^c, I^s_i  \right)\right\},
\end{equation}
where $M_i^s$, $y_i^c$ and $I^s_i$ denote the cropped binary mask of the $i$-th object, its category and its corresponding RGB segment image, respectively. 
Besides, for background region without instance-level annotation, we take the maximal connected component as a single background object. 
Decomposing all the images in training dataset, we create a retrieval database, which is used in both training and testing stage. 


\noindent \textbf{Retrieval Strategy.}
Given a semantic map $M$, we first decompose it into a number of segment masks $\{(M_i^s, y_i^c)\}$. 
%
Then, we retrieve the most compatible segment from the retrieval database for each segment mask. 
Specifically, for segment mask $M^s_i$ with category $y^c_i$, we retrieve a segment $(M_j^s, y_j^c, I^s_j)$ which has the same category ($ y_j^c =  y_i^c$) and similar shape with $M^s_i$. 
To measure the similarity between two segment masks ($M^s_i$ and $M^s_j$), we adopt the geometric score~\cite{wang2020constrained} to measure both scale and shape consistency,
\begin{equation}
\begin{split}
    \sigma_{scale}\left(M^s_i,M^s_j\right) = \begin{cases} 0, & t \ge 0.5 \\ 1, & t < 0.5 \end{cases}, \\
\end{split}
\end{equation}
\begin{equation}
    \sigma_{shape}\left(M^s_i,M^s_j\right) = \frac{SSD\left(\hat{M}^s_i, \hat{M}^s_j\right)}{\max\left( \left \Vert \hat{M}^s_i \right \Vert_1, \left \Vert \hat{M}^s_j \right \Vert_1 \right)},
\end{equation}
where $ t = \frac { \min \left ( \left \Vert  M^s_i \right \Vert_1, \left \Vert M^s_j \right  \Vert_1 \right ) } { \max \left ( \left \Vert  M^s_i \right \Vert_1, \left \Vert M^s_j \right  \Vert_1 \right ) } $. 
$\hat{M}^s_i$ and $\hat{M}^s_j$ denote the resized versions (\ie, $128\times 128$) of ${M^s_i}$ and ${M^s_j}$ using nearest neighbor interpolation, respectively. 
$SSD\left(\cdot\right)$ denotes the sum square difference. 
The final consistency is calculated as,
\begin{equation}
\small
    \sigma\left(M^s_i, M^s_j\right) = \sigma_{scale}\left(M^s_i,M^s_j\right) + \gamma \sigma_{shape}\left(M^s_i,M^s_j\right).
    \label{eq:non-similarity}
\end{equation}
where $\gamma$ is the balance coefficient and we set $\gamma=1$ in practice. Lower $\sigma\left(M^s_i, M^s_j\right)$ indicates more similarity between two segment masks.


\noindent \textbf{Composition of Guidance Image.} 
Finally, we recompose the retrieved segments as the guidance image. Let  $(M_r^s,y_r^c,I_r^s)$ denotes the retrieved segment for the given segment mask $M_i^s$. As illustrated in Fig.~\ref{fig:composition}, $I_r^s$ and the corresponding mask ${M_r^s}$ are first resized to the size of $M_i^s$. The resized mask and image are denoted as $\hat{M}_r^s$ and $\hat{I}_r^s$. Then, the resized image is pasted into the guidance image according to the original position of $M_i^s$.  To maintain integrity of instance, we paste the segment image following the below rules:
\begin{itemize}
    \setlength{\itemsep}{0pt}
	\setlength{\parsep}{0pt}
	\setlength{\parskip}{0pt}
    \item Pixels of $\hat{I}^{s}_r$ in both $\hat{M}_r^s$ and $M_i^s$ are preserved.
    \item If $ y_r^c $ belongs to background things categories, pixels of $\hat{I}^{s}_r$ in $\hat{M}_r^s$ but not in $M_i^s$ are zeroed out.
    \item If $ y_r^c $ belongs to foreground (\ie, instance object) and pixels of $\hat{I}^{s}_r$ in $\hat{M}_r^s$ but not in $M_i^s$ are located in the \textit{background} categories in $M$, they are preserved.
    \item If $ y_r^c$ belongs to foreground and pixels of $\hat{I}^{s}_r$ in  $\hat{M}_r^s$ but not in $M_i^s$ are located in the \textit{foreground} categories in $M$, they are zeroed out.
\end{itemize}

We finally obtain the retrieval-based guidance image $I^r$ to guide the image synthesis.



\subsection{Distortion of Ground-truth Image.} 
To distort the ground-truth image $I^{gt}$, we first decompose it into a set of segment images $I^{gt} = \{I^{s}_i\}$. 
Then we apply the distortion (\ie, color, shape and resolution) on each segment image $I^s_i$.

\noindent \textbf{Color.} 
We employ the method proposed by~\cite{reinhard2001color} to transfer the color of segment image $I^s_i$ to a random segment image $I^s_t$ with the same category. 
Specifically, we first convert $I^s_i$ and $I^s_t$ from $RGB$ space into $l \alpha \beta$ space. 
Then the color transferred image ${\tilde{I}^s_i}$ in each channel of $l \alpha \beta$ space is calculated by,
\begin{equation}
\begin{split}
        \tilde{l}_i &= \left( l_i -\mu(l_i) \right) \cdot \frac{\sigma(l_t) }{\sigma(l_i)}  + \mu(l_t)\\
        \tilde{\alpha}_i &= \left( \alpha_i -\mu(\alpha_i)\right) \cdot \frac{\sigma(\alpha_t)}{\sigma(\alpha_i)}  + \mu(\alpha_t) \\
        \tilde{\beta}_i &= \left( \beta_i -\mu(\beta_i)\right) \cdot \frac{\sigma(\beta_t)}{\sigma(\beta_i)}  + \mu(\beta_t) \\
\end{split}
\end{equation}
where $\mu(\cdot)$ and $\sigma(\cdot)$ denote the mean and standard deviation of corresponding channel. 
Finally, we convert ${\tilde{I}^s_i}$ from $l \alpha \beta$ into $RGB$ space to obtain the color distorted image. 

\noindent \textbf{Shape.} 
To distort the shape of a segment image, we first sample 10 points uniformly on the edge of the segment image as source points, and shift three of them randomly to produce the target points. 
The source points and target points are used to produce a dense flow utilizing thin plate spline algorithm. 
Then we use the produced flow to warp the segment image to obtain the shape distorted image.

\begin{figure*}
	\centering
	\includegraphics[width=0.99 \linewidth]{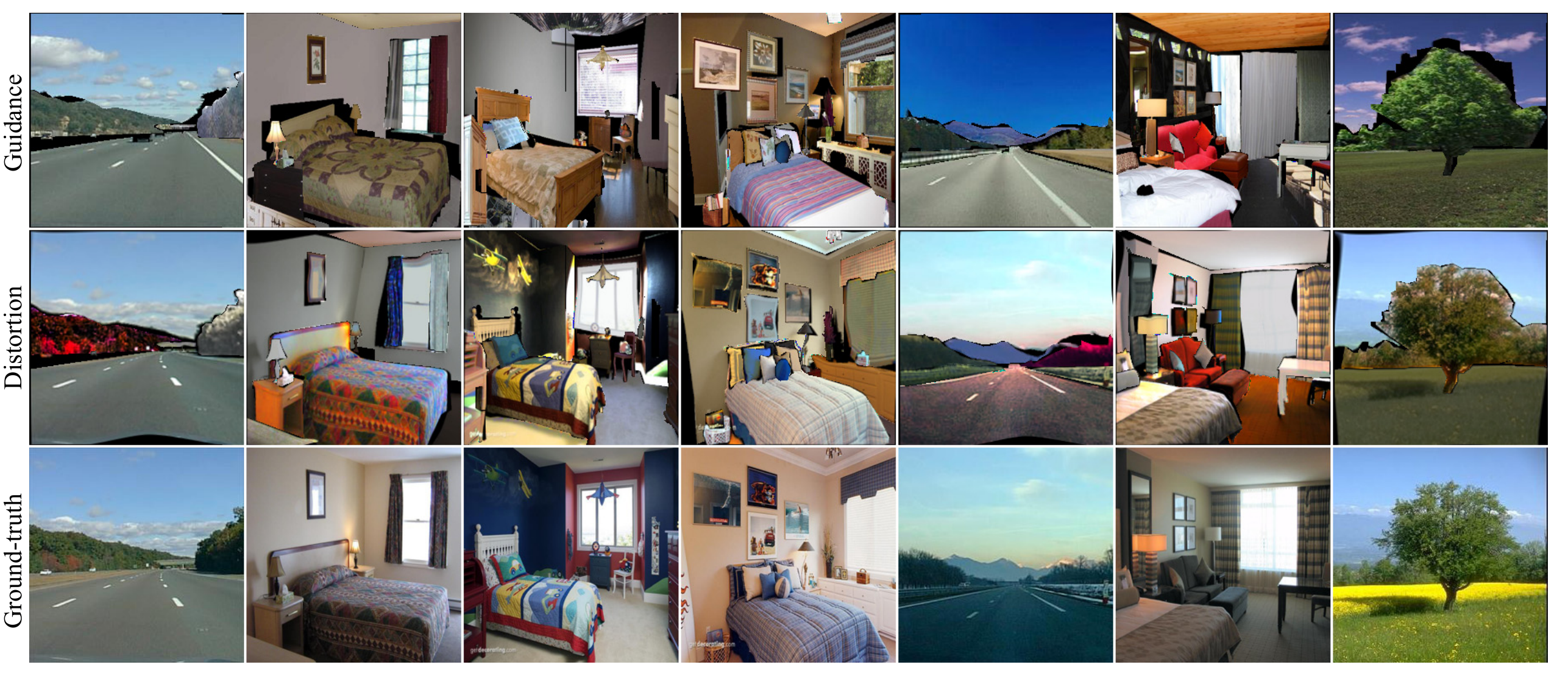}
	\caption{Distortion of ground-truth images. The top row shows the produced retrieval-based guidance images; the middle row shows the distorted ground-truth and the bottom row shows the corresponding ground-truth images.}
	\label{fig:dist}
	\vspace{-0.8em}
\end{figure*}

\noindent \textbf{Resolution.} To distort the resolution of a segment image, we downsample it with a random scale $\tau\space (0.5<\tau<1)$, and upsample it to the original size.

After distortion, distorted segment images from ground-truth $I^{gt}$ recompose the distorted ground-truth $\Tilde{I}^{gt}$ to facilitate model training. The distortion results are shown in Fig.~\ref{fig:dist}.

\section{Additional Details of Training Architecture}

\noindent \textbf{Details of RESAIL module.}
The RESAIL module takes both the guidance image (\ie, retrieval-based guidance $I^r$ or distorted ground-truth $\Tilde{I}^{gt}$) and the semantic map $M$ as input and learns to modulate the activations. We here represent the input activations as $\mathbf{h}$  with a batch of $N$ samples. $H$, $W$ and $C$ denote the height, width and the number of channels in $\mathbf{h}$, and the modulated activations at site~$(n\in N, c\in C, y \in H, x \in W)$ is represented as,

\begin{equation}
\begin{split}
        RESAIL(\mathbf{h}, I^r, M) &= \gamma_{c, y, x}\left(I^r, M\right) \frac{\mathbf{h}_{n,c,y,x} - \mu_c}{\sigma_c} \\ 
        &+ \beta_{c,y,x}\left(I^r, M\right), \\
\end{split}
\label{eq:mod}
\end{equation}
where $\mu_c$ and $\sigma_c$  denote the mean and standard deviation of the activation in channel $c$,
\begin{equation}
    \begin{split}
      \mu_c &= \frac 1 {NHW}\sum_{n, y, x}\mathbf{h}_{n, y, x} \\ \sigma_c &= \sqrt{\frac 1 {NHW} \left( \sum_{n, y, x}\mathbf{h}_{n, y, x}^2 \right) -\mu_c^2} \\
    \end{split}.
\end{equation}
$\gamma(\cdot)$ and $\beta(\cdot)$ 
have the same architectures and learn the parameters for modulating the scales and biases, respectively. 
We here take $\gamma(\cdot)$ as an example, which consists of two separated convolutional neural networks to produce coarse and fine-grained guidance for modulation. 
The one network $\gamma^s(\cdot)$ takes the semantic map $M$ to learn the coarse modulation parameters.
The other network $\gamma^r(\cdot)$ takes the retrieved image $I^r$ to learn the pixel-level fine-grained modulation parameters, and we also take the semantic map $M$ to modulate the intermediate features with AdaIN blocks.
\begin{equation}
    \footnotesize
    \hspace{-2em}
    \begin{split}
        \gamma_{c, y, x}\left(I^r, M\right) = \alpha_{\gamma}\gamma_{c, y, x}^s\left( M\right) + \left(1 - \alpha_{\gamma} \right) \gamma_{c, y, x}^r\left(I^r, M\right) \\
        \beta_{c,y,x}\left(I^r, M\right) = \alpha_{\beta}\beta_{c, y, x}^s\left( M\right) + \left(1 - \alpha_{\beta} \right) \beta_{c, y, x}^r\left(I^r, M\right) \\
    \end{split},
\end{equation}
where the $0<\alpha_{\beta}, \alpha_{\gamma}<1$ are learnable scalars.

\begin{figure}
\vspace{-0.8em}
	\centering
	\includegraphics[width=0.99 \linewidth]{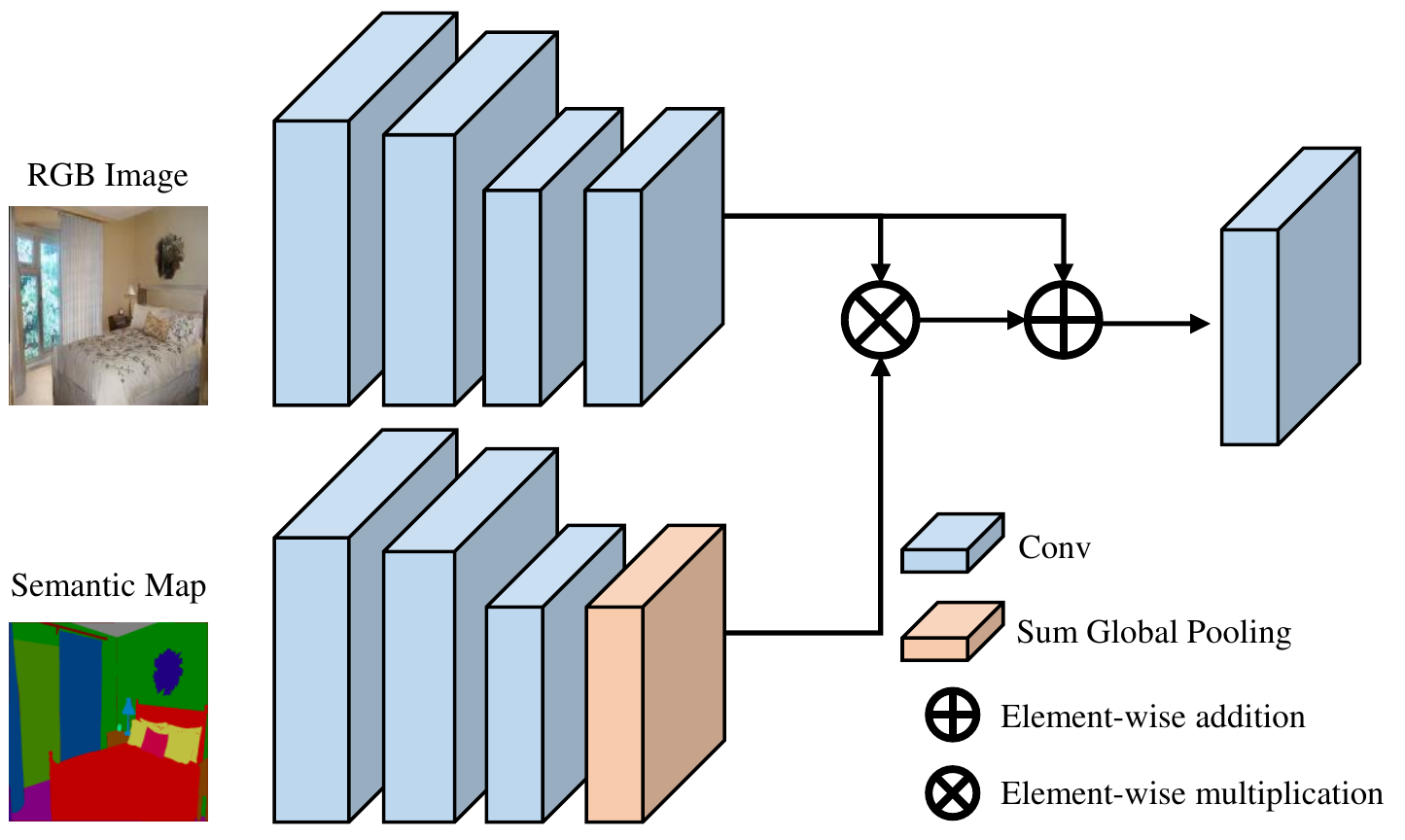}
	\caption{Discriminator network. }
	\label{fig:net-D}
	\vspace{-0.8em}
\end{figure}

\noindent \textbf{Discriminator.}
In practice, we adopt two multi-scale discriminators proposed by~\cite{ntavelis2020sesame} to facilitate our model training.
As shown in Fig.~\ref{fig:net-D}, the discriminator consists of two pathways and processes the RGB image and the semantic labels respectively; then the final features are merged by element-wise addition and element-wise multiplication. 

\section{Additional Ablation Studies}

\begin{figure*}
	\centering
	\includegraphics[width=0.95 \linewidth]{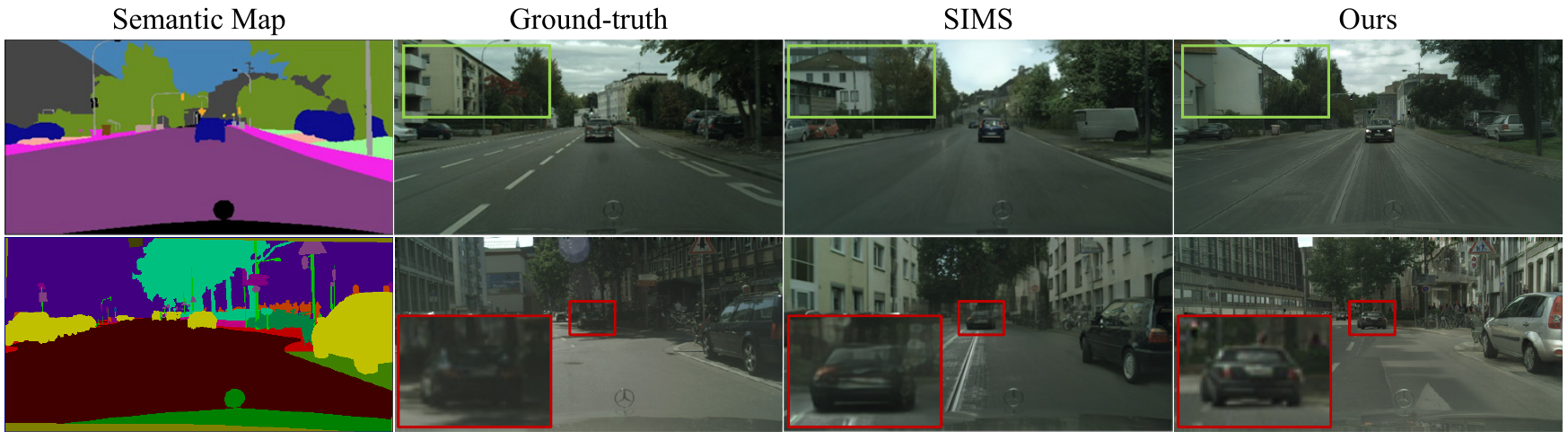}
	\caption{Comparison with SIMS. SIMS suffers from low mIOU (marked in green rectangle) and blurs (marked in red rectangle) of some objects.}
	\label{fig:cmp_sims}
	\vspace{-0.8em}
\end{figure*}

\noindent \textbf{Comparison with SIMS.} Also introducing an image synthesis mechanism based on reference, SIMS~\cite{qi2018semi} simply takes the retrieved image as  network input, resulting in low mIOU and blurs shown as Fig.~\ref{fig:cmp_sims} and Table~{\color{red}{1}}. While our method leverages the retrieved images to provide pixel level fine-grained guidance via spatially adaptive normalization, making it more effective in synthesizing photo-realistic images.

\begin{figure*}
	\centering
	\includegraphics[width=0.99 \linewidth]{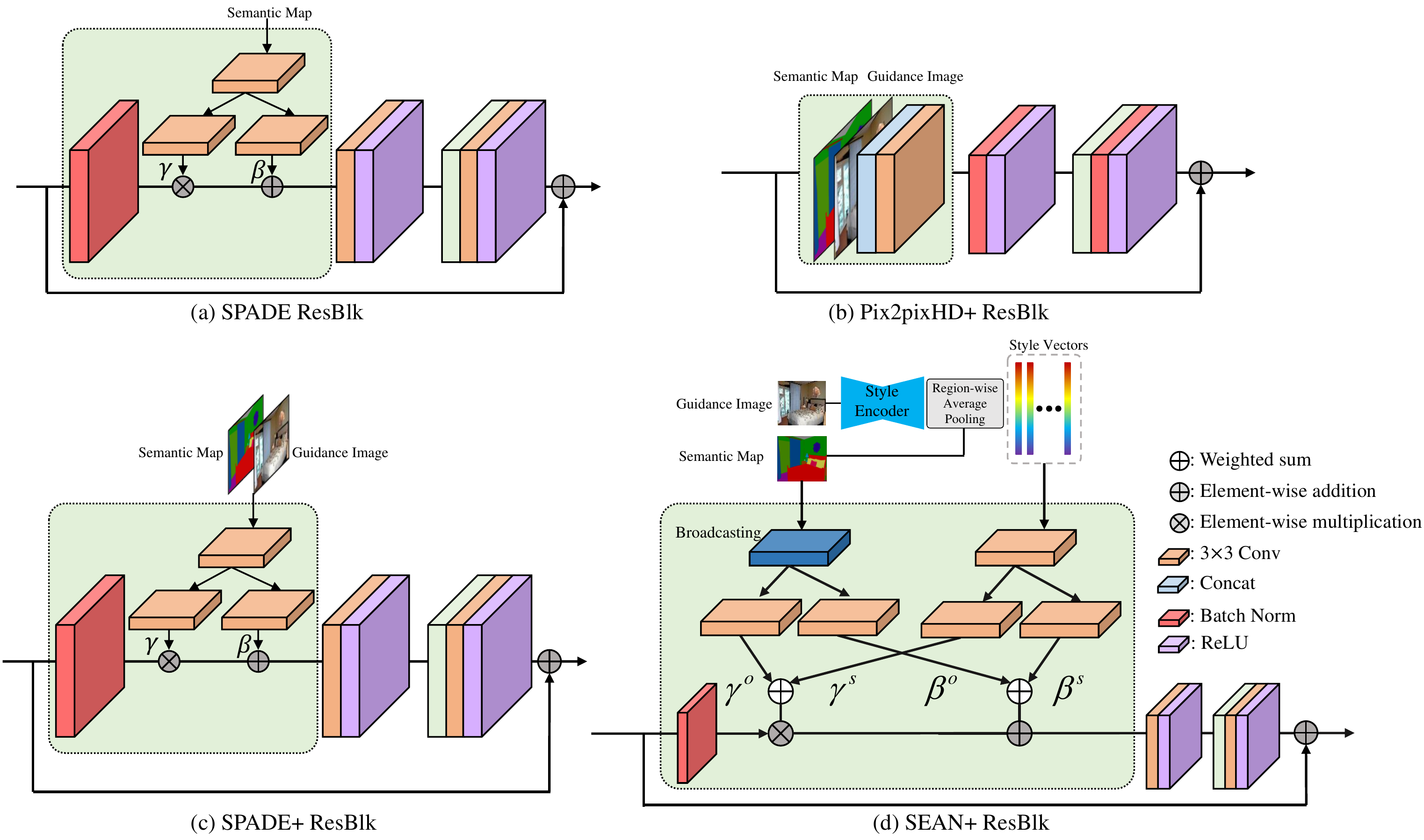}
	\caption{Variants of RESAIL ResBlk. (a) \textit{SPADE} employs the SPADE module; (b) \textit{Pix2pixHD+} denotes concatenating the guidance into the conv layer of pix2pixHD model. (c) \textit{SPADE+} denotes using the guidance as input to the SPADE module. (d) \textit{SEAN+} denotes using the guidance as input to the SEAN module.}
	\label{fig:struc}
	\vspace{-0.9em}
\end{figure*}

\noindent \textbf{Variants of RESAIL.}
We compare our RESAIL module with 4 variants and in each comparison experiment we employ the same generator architecture while only replacing the RESAIL ResBlk with other variants. 
We show the different ResBlks in Fig.~\ref{fig:struc}. 
In \textit{SPADE}, we just employ the module proposed by~\cite{park2019semantic}.
In \textit{SPADE+}, semantic map concatenating with the guidance image is convolved to produce the modulation parameters $\beta$ and $\gamma$. 
In \textit{Pix2pixHD+}, we concatenate the feature with the semantic map and the guidance image following with convolution layer, and we discard the encoder part of Pix2pixHD~\cite{wang2018high}. 
In \textit{SEAN+}, we extract per region style vectors from the guidance image with a style encoder network and input the style vector and semantic map into the SEAN~\cite{zhu2020sean} module. Limited by GPU memory, dimension of style vector is set to 128.

\begin{figure*}
	\centering
	\includegraphics[width=0.90 \linewidth]{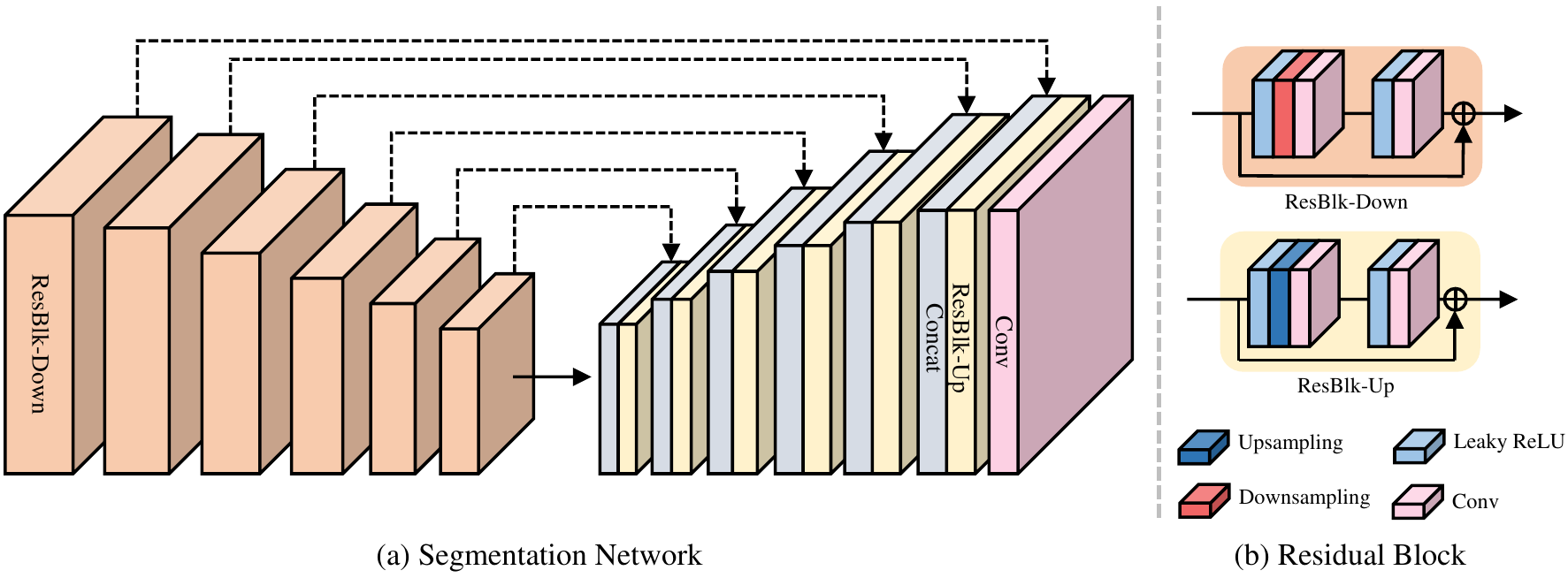}
	\caption{Segmentation network. (a) The network is designed based on U-Net. (b) Each downsampling or upsampling operation employs a ResBlk.  }
	\label{fig:u-net}
\end{figure*}

\begin{table}[t]
	\small{
		\begin{center}
			\caption{Ablation study of $\mathcal{L}_{seg}$ in Cityscapes dataset. It shows that $\mathcal{L}_{seg}$ facilitates the model learning.}
			\label{tab:seg_loss_aba}
			
			\resizebox{0.6\linewidth}{!}{
				\aboverulesep=0ex
				\belowrulesep=0ex
				\renewcommand{\arraystretch}{1.2}
				\begin{tabular}{cccc}
					\toprule
					$\mathcal{L}_{seg}$ & FID$\left(\downarrow\right)$ & mIOU$\left(\uparrow\right)$ & AC$\left(\uparrow\right)$  \\
					\midrule
					
					\xmark & 46.8 & 66.3 & 82.7 \\
					
					\cmark & \textbf{45.5} & \textbf{69.7} & \textbf{83.2} \\
					\bottomrule
					\vspace{0.1pt}
				\end{tabular}
			}
			
		\end{center}
	}
	\vspace{-0.2in}
\end{table}

\begin{figure*}
	\centering
	\includegraphics[width=0.99 \linewidth]{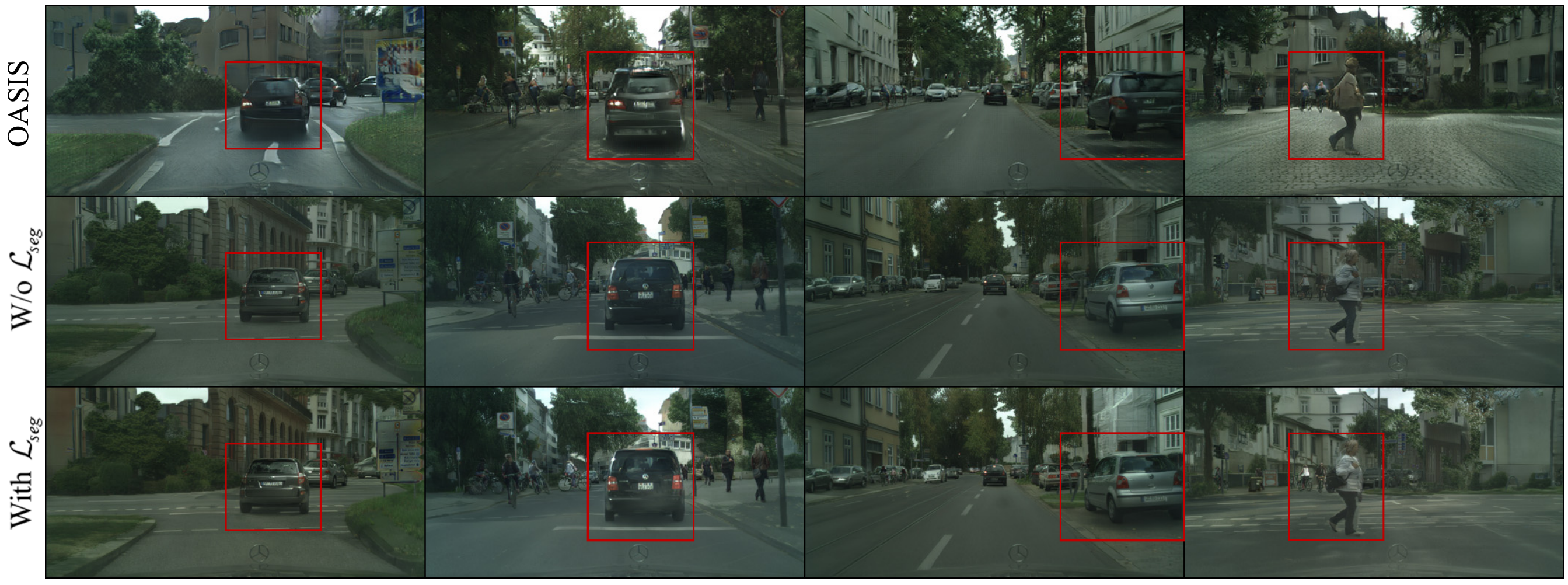}
	\caption{Effect of segmentation loss $\mathcal{L}_{seg}$. Red rectangles mark the affected instances. OASIS suffers from inconsistent edge transitions whose discriminator based on a segmentation network. With the help of  other losses (\eg, GAN loss and perceptual loss), no obvious edge transitions are found in our results with $\mathcal{L}_{seg}$.}
	\label{fig:edge}
\end{figure*}

\noindent \textbf{Effectiveness of $\mathcal{L}_{seg}$.}
To prompt the model to synthesize images aligning well with the semantic layout, we introduce a pretrained segmentation network $ C $ to classify each pixel of the generated image and optimize the segmentation loss $\mathcal{L}_{seg}$. The designed segmentation network $ C $ follows~\cite{schonfeld2021you}, which consists of 12 ResBlks based on a U-Net architecture as shown in Fig.~\ref{fig:u-net}. We report the results of training our model with and without $\mathcal{L}_{seg}$ on Cityscapes~\cite{cordts2016cityscapes} in Table~\ref{tab:seg_loss_aba}. From the table, we can see segmentation loss $\mathcal{L}_{seg}$ improves the learning process. Albeit $\mathcal{L}_{seg}$ helps segmentation based metrics, it may introduce inconsistent edge transitions among instances, occurring in \cite{schonfeld2021you} which introduces a discriminator based on a segmentation network shown as Fig.~\ref{fig:edge}. However, with other losses (\eg, GAN loss and perceptual loss) prompting model training, this kind of artifacts are suppressed and no obvious transitions are found in our results with $\mathcal{L}_{seg}$.

\begin{table}[t]
	{\small
		\begin{center}
		
		\caption{FID w.r.t non-similarity threshold.}
	    \vspace{-0.9em}
	   \label{tab:fid_thres}
		\resizebox{0.90\linewidth}{!}{
			\aboverulesep=0ex
			\belowrulesep=0ex
			\renewcommand{\arraystretch}{1.2}
			\begin{tabular}{rcccccccccc}
				\toprule
				Threshold & 0.15 & 0.25 & 0.35 & 0.45 & 0.55 & 0.58 \\
				FID & 45.49 & 46.38 & 48.18 & 48.3 & 50.56 & 51.04 \\
				\bottomrule
			\end{tabular}
		}
		\end{center}
	}
\vspace{-0.9em}

\end{table}

\noindent \textbf{Effect of Shape Non-similarity Threshold.}
Computed as Eq.~\ref{eq:non-similarity}, non-similarity $\sigma$ is adopted to measure the shape consistency between two segment masks. We have tested the FID results by adopting different non-similarity thresholds. From Table~\ref{tab:fid_thres}, higher threshold (\ie, using more non-similar guidance) leads to worse guidance, resulting in worse FID.

\vspace{-0.8em}
\section{Additional Visual Results}
\begin{figure*}
	\centering
	\includegraphics[width=0.99 \linewidth]{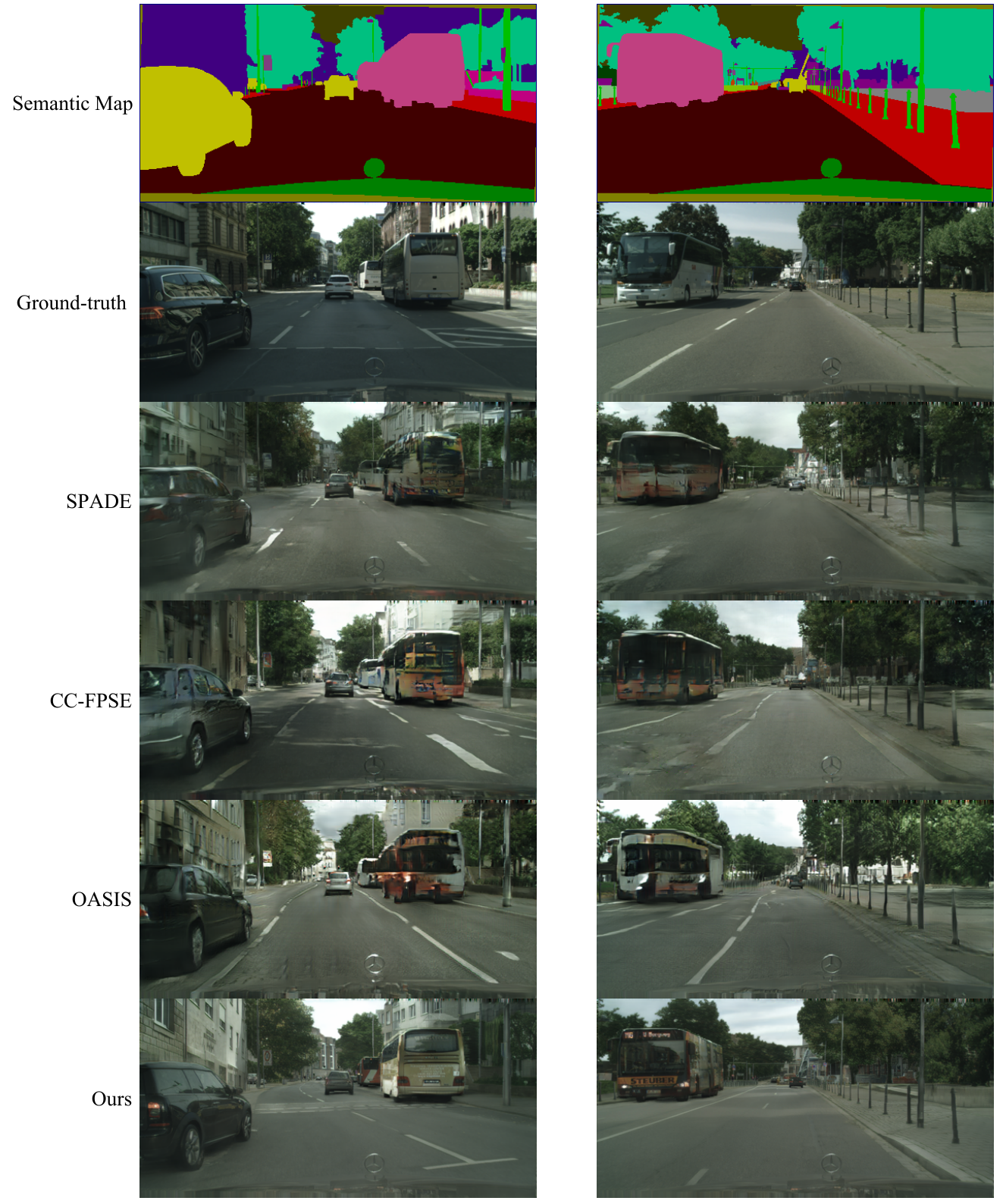}
	\caption{Comparison results on Cityscapes.}
	\label{fig:city1}
\end{figure*}

\begin{figure*}
	\centering
	\includegraphics[width=0.99 \linewidth]{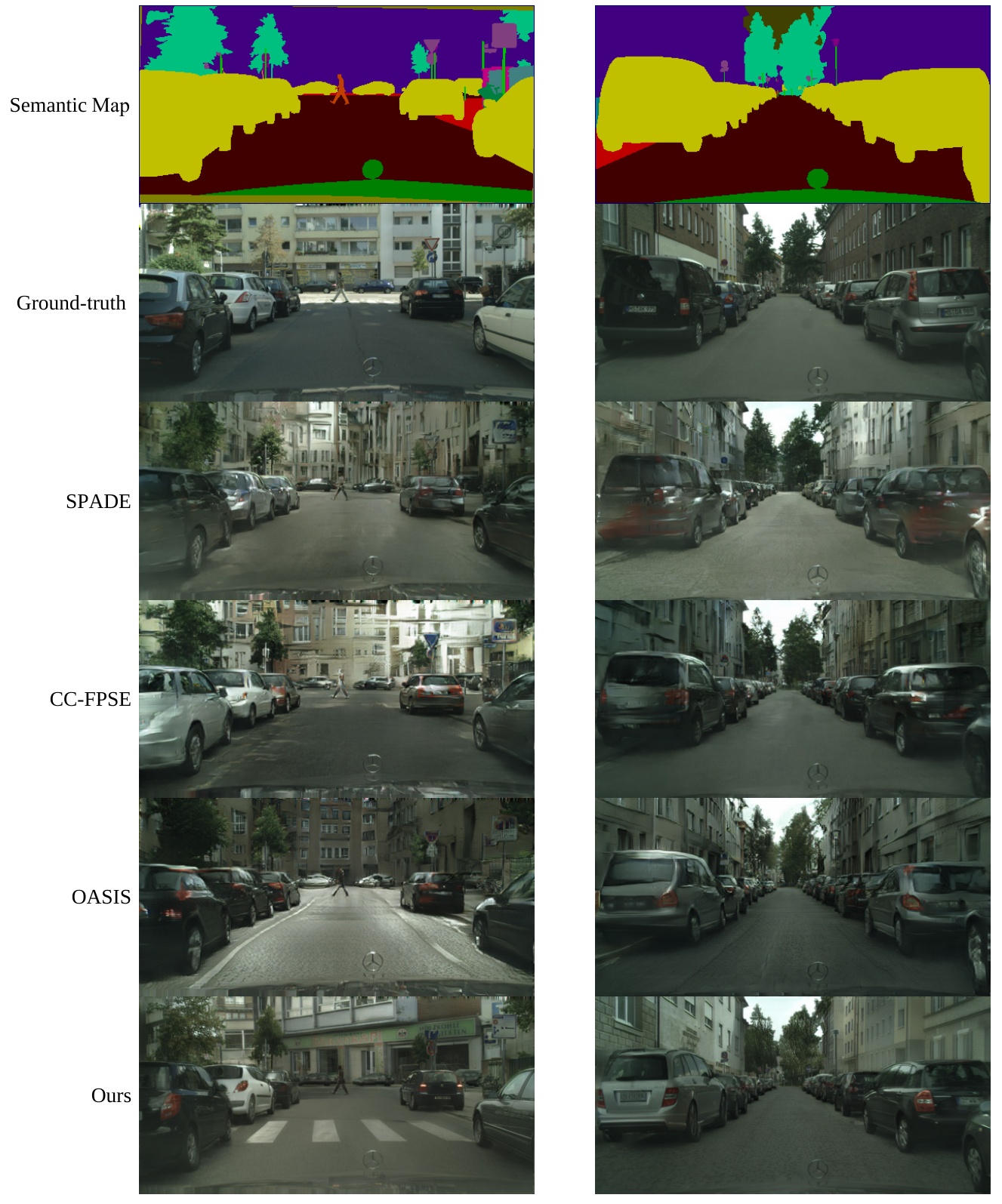}
	\caption{Comparison results on Cityscapes.}
	\label{fig:city3}
	\vspace{5em}
\end{figure*}

\begin{figure*}
	\centering
	\includegraphics[width=0.99 \linewidth]{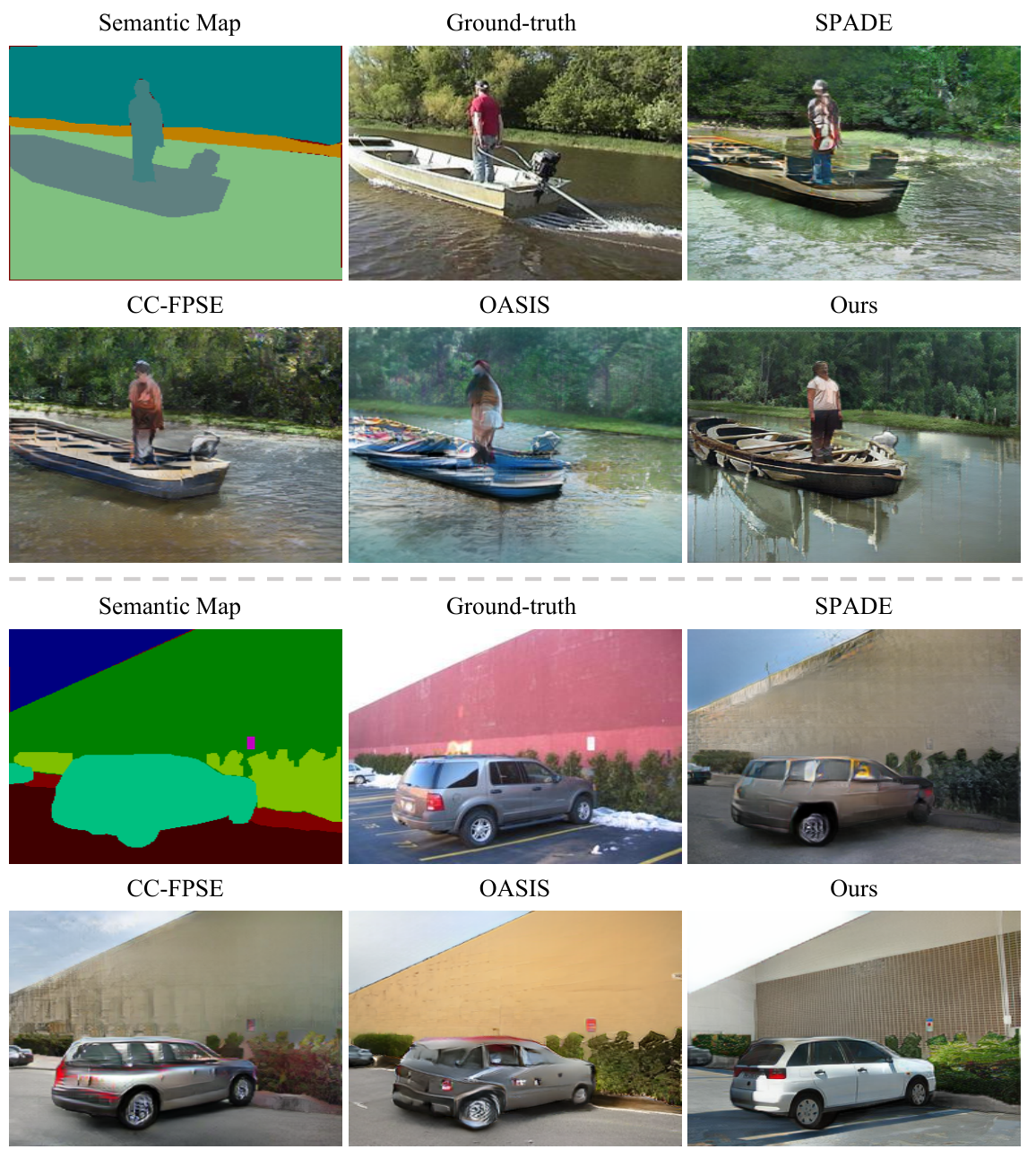}
	\caption{Comparison results on ADE20K.}
	\label{fig:ade1}
	\vspace{5em}
\end{figure*}

\begin{figure*}
	\centering
	\includegraphics[width=0.99 \linewidth]{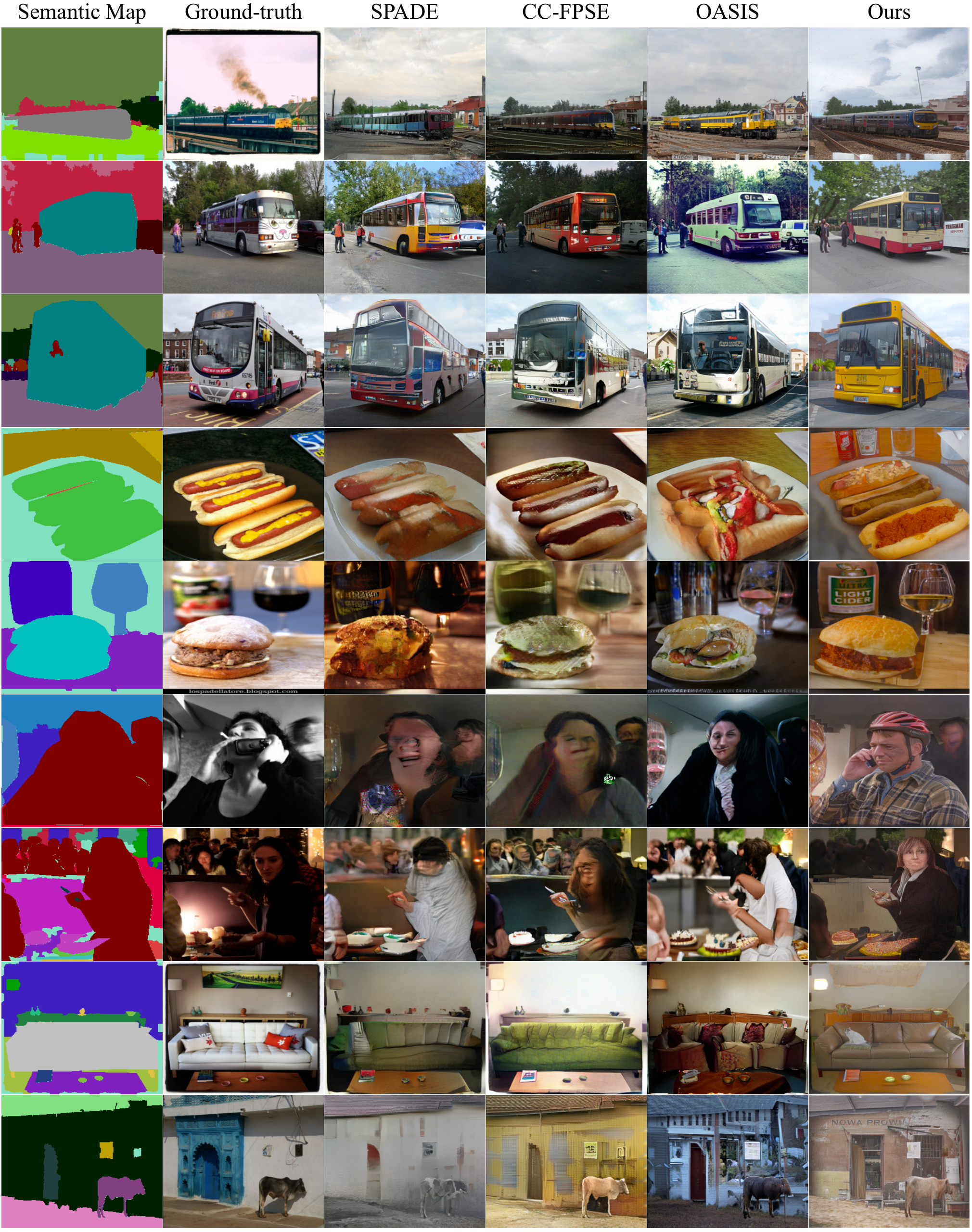}
	\caption{Comparison results on COCO-Stuff.}
	\label{fig:coco_cmp}
\end{figure*}

\begin{figure*}
	\centering
	\includegraphics[width=0.99 \linewidth]{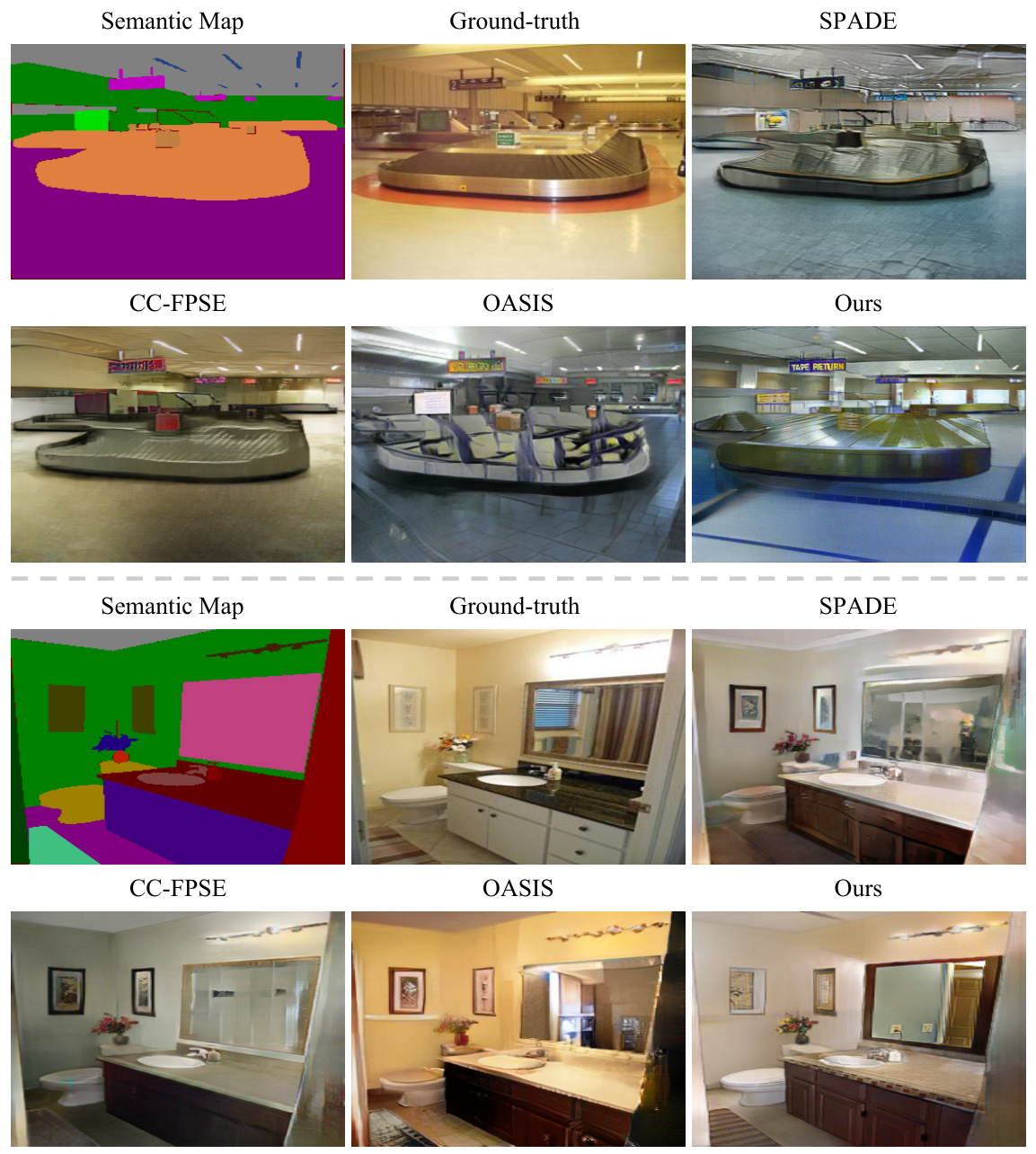}
	\caption{Comparison results on ADE20K.}
	\label{fig:ade2}
	\vspace{15em}
\end{figure*}

\begin{figure*}
	\centering
	\includegraphics[width=0.99 \linewidth]{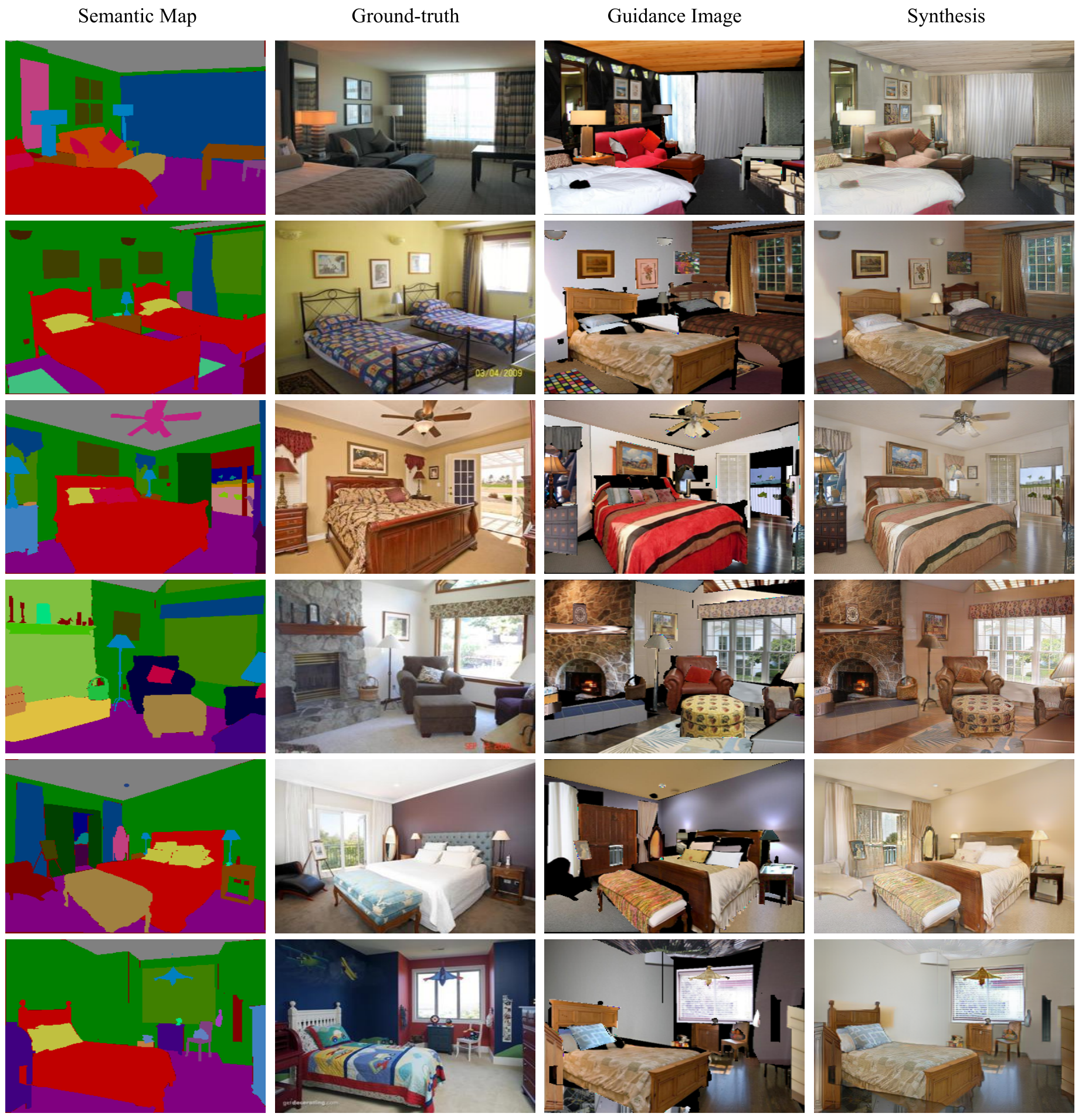}
	\caption{Synthesis results on ADE20K.}
	\label{fig:aderesults}
	\vspace{15em}
\end{figure*}

\begin{figure*}
	\centering
	\includegraphics[width=0.99 \linewidth]{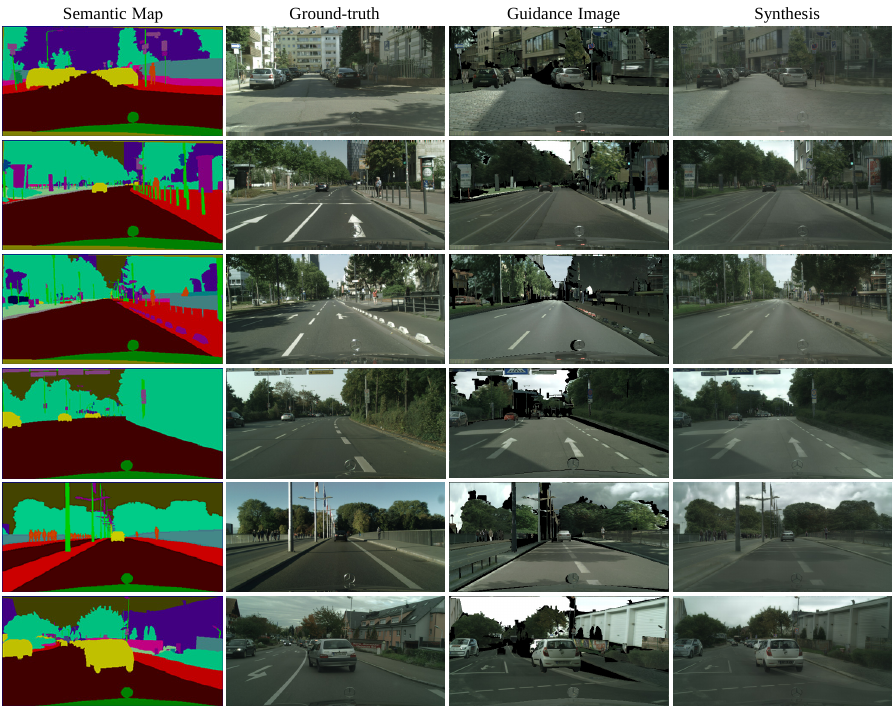}
	\caption{Synthesis results on Cityscapes.}
	\label{fig:cityresults}
	\vspace{20em}
\end{figure*}

\vspace{-0.8em}
To demonstrate the effectiveness of our method on synthesizing the photo-realistic images, we show more visual results in this section. Fig.~\ref{fig:city1} $\sim$ \ref{fig:city3} show the comparisons on Cityscapes\cite{cordts2016cityscapes} and as shown in figures, our synthesized images are more photo-realistic with fine details. Fig.~\ref{fig:ade1} and Fig.~\ref{fig:ade2} show more results on ADE20K~\cite{zhou2017scene}. Comparisons on COCO-Stuff~\cite{caesar2018coco} can be found in Fig.~\ref{fig:coco_cmp}. The guidance image and its corresponding generated image are shown as Fig.~\ref{fig:aderesults} and Fig.~\ref{fig:cityresults}.

\begin{figure*}
	\centering
	\includegraphics[width=0.99 \linewidth]{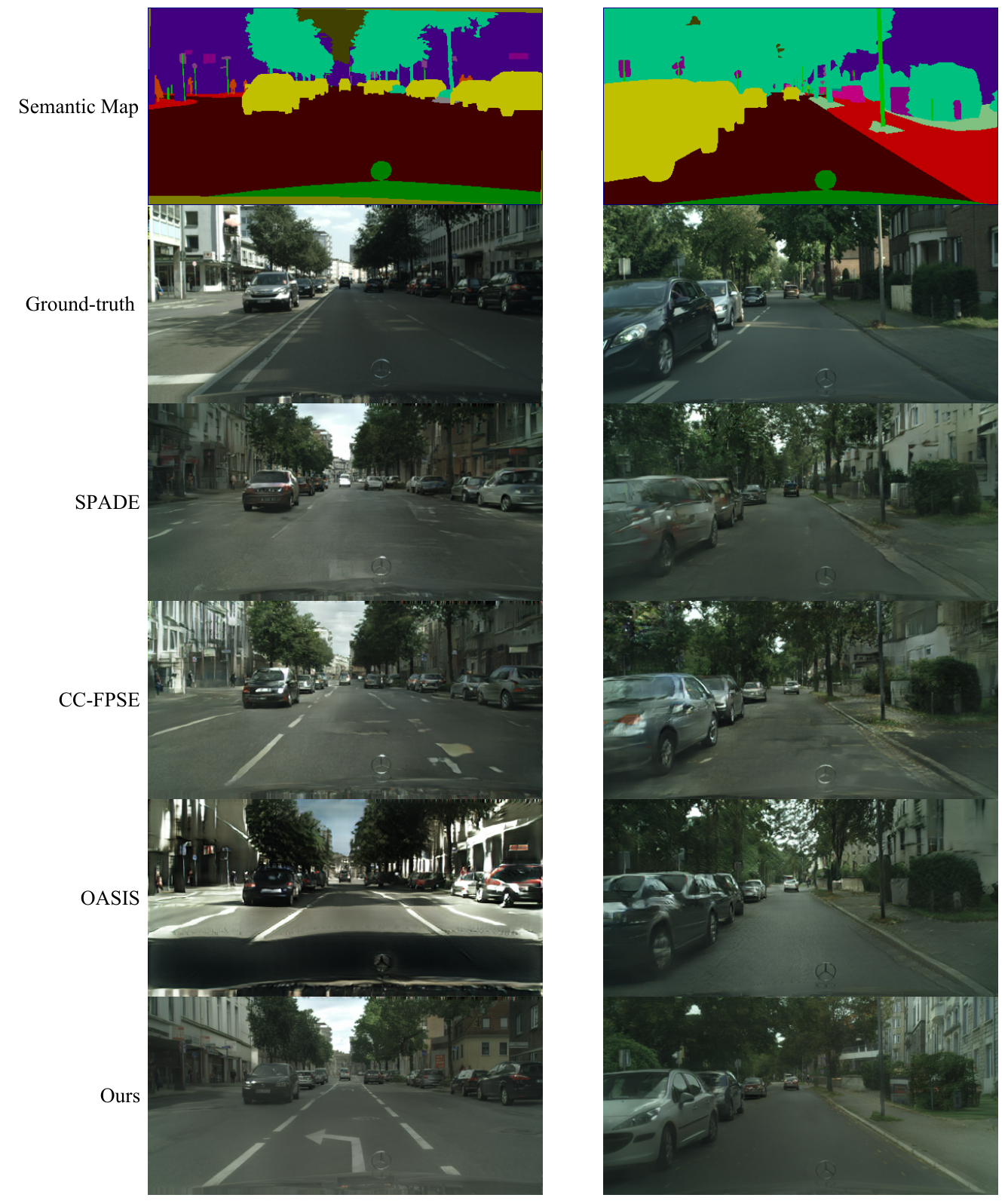}
	\caption{Comparison results on Cityscapes.}
	\label{fig:city2}
\end{figure*}

\begin{figure*}
	\centering
	\includegraphics[width=0.99 \linewidth]{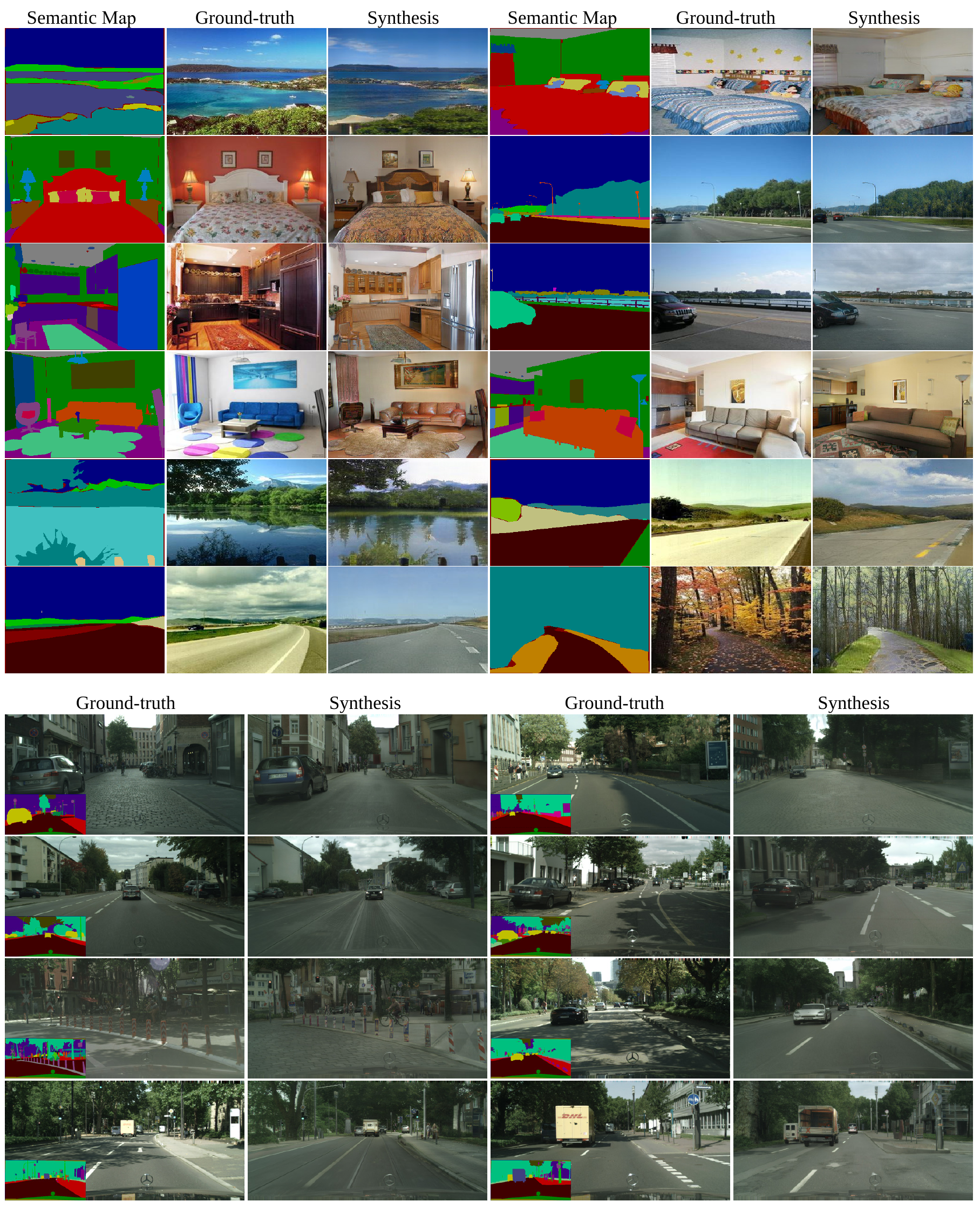}
	\caption{Synthesis results on ADE20K(top) and Cityscapes(bottom).}
	\label{fig:mores}
\end{figure*}

\end{document}